\documentclass[11pt,a4paper]{article}
\usepackage[hyperref]{acl2019}
\usepackage{times}
\usepackage{latexsym}
 
\usepackage{verbatim}
\usepackage{xspace} 
\usepackage{graphicx}
\usepackage{adjustbox}
\usepackage{array}
\usepackage{multirow}
\usepackage{booktabs}
\usepackage{caption}
\usepackage{subcaption}

\newcounter{trCounter}
\newif\iftrvar
\trvartrue
\iftrvar
\newcommand{\tim}[1]{{\small \color{blue} \refstepcounter{trCounter}\textsf{[TR]$_{\arabic{trCounter}}$:{#1}}}}
\else
\newcommand{\tim}[1]{}
\fi

\newcommand{\numlocations}{663 }
\newcommand{\numcharacters}{1755 }
\newcommand{\numobjects}{3462 }
\newcommand{\numcategories}{37 }

\usepackage{url}

\aclfinalcopy 

\usepackage{txfonts}
\newcommand{\fair}{$^1$}
\newcommand{\ucl}{$^3$}
\newcommand{\loria}{$^2$}
\newcommand{\uclfair}{$^{1,3}$}
\newcommand{\loriafair}{$^{1,2}$}

\title{Learning to Speak and Act in a Fantasy Text Adventure Game}

\author{
Jack Urbanek\fair{} \ Angela Fan\loriafair{} \ Siddharth Karamcheti\fair{} \ Saachi Jain\fair{} \ Samuel Humeau\fair{} \\
{\bf Emily Dinan\fair{} \ Tim Rockt\"aschel\uclfair{}  \ Douwe Kiela\fair{} \ Arthur Szlam\fair{} \ Jason Weston\fair{}}\\
\fair{}Facebook AI Research \\
\loria{}LORIA, Nancy\\
\ucl{}University College London\\
{\tt light-dms@fb.com}
}
 
\date{}

\begin{document}
\maketitle
\begin{abstract}
We introduce a large-scale crowdsourced 
text adventure game as a research platform for studying grounded dialogue. 
In it, agents can perceive, emote, and act whilst 
conducting dialogue with other agents. 
Models and humans can both act as characters within the game.
We describe the results of training state-of-the-art generative and retrieval models in this setting.  We show that in addition to using past dialogue, these models are able to effectively use the state of the underlying world to condition their predictions.
 In particular, we show that  grounding on the details of the local environment, including location descriptions, and the objects 
(and their affordances) and characters (and their previous actions) present within it allows better predictions of agent behavior and dialogue. 
We analyze the ingredients necessary for successful grounding in this setting,
and how each of these factors relate to agents that can talk
and act successfully. 
\end{abstract}

\section{Introduction}

There has been remarkable progress in language modeling \cite{jozefowicz2016exploring, devlin2018bert, gpt2} and building dialogue agents \cite{dinan2019second}.  Nevertheless, the current state of the art  
uses only the statistical regularities of language data, without explicit understanding of the world that the language describes. 
This work is built on the hypothesis that dialogue agents embodied in a rich and cohesive (but tractable) world can more easily be trained to use language effectively than those only exposed to standard large-scale text-only corpora.

To that end, we introduce the LIGHT\footnote{
Learning in Interactive Games with Humans and Text.} research platform.   LIGHT is a multi-player fantasy text adventure world designed for studying situated dialogue, and allows
interactions between humans, models as embodied agents, and the world itself.  It consists of a large crowdsourced game world (\numlocations locations, \numobjects objects and \numcharacters characters) described entirely in natural language.
Within that game world, we collect a large set (11k episodes)
of character-driven human-human crowdworker interactions involving
actions, emotes, and dialogue,
with the aim of training models to engage humans in a similar fashion.
Our framework is made publicly available in ParlAI ({\small{\url{http://parl.ai/projects/light}}}).

We use the collected dataset to investigate how a model can both speak {\em and} act grounded in perception of its environment and dialogue from other speakers. This is done by evaluating state-of-the-art models on our task and evaluating the effects of providing additional grounding. In particular, we adapt the BERT 
contextual language model \citep{devlin2018bert}
to the task of dialogue in two ways: as 
a bi-ranker, which is fast and practical as a retrieval model,
and as a cross-ranker which is slower at inference time but allows more
feature cross-correlation between context and response. Both models outperform existing methods. Our ablation analysis shows the importance of each part of the grounding (location, objects, characters, other's actions, self-actions) in terms of the ability to both understand and  use language. 
While models that use grounding show clear improvements, our best performing models are still unable to perform at human level,  making our setup a suitable challenge for future research.


\section{Related Work}

Most recent work in dialogue exploring generative or retrieval models for goal-directed \citep{henderson2014second,bordes2016learning} 
or chit-chat tasks \citep{vinyals2015neural,sordoni2015neural,zhang2018personalizing} is not situated, or even grounded in perception.
Models typically take the last few utterances from 
the dialogue history as input, and output a new utterance.
While some goal-directed setups may use external knowledge bases (e.g. flight data for airline booking), dialogues tend to implicitly refer to an external world during the conversations without explicit grounding to objects or actions.

Several position papers have proposed virtual embodiment as a strategy for language research
\cite{brooks1991intelligence,kiela2016virtual,gauthier2016paradigm,mikolov2016roadmap,lake2017building}. Single-player text adventure game frameworks for training reinforcement learning agents exist, i.e., \newcite{narasimhan2015language} and TextWorld \cite{cote2018textworld}, but these do not have human dialogue within the game. \newcite{yang2017mastering} and \newcite{bordes2010towards} proposed small world setups for instruction following or labeling, but these are much more restricted than the large multi-player text adventure game environment with rich dialogue that we propose here.

A number of visual, rather than text, platforms have been proposed such
as  House3D \cite{wu2018building}, HoME \cite{brodeur2017home}, MINOS \cite{savva2017minos}, Matterport3D \cite{chang2017matterport3d} and AI2-THOR \cite{kolve2017ai2},
and the Minecraft MALMO project \cite{johnson2016malmo}, 
but they typically are suited to reinforcement learning of actions, and 
involve templated language for navigation or question answering tasks, if at all \cite{oh2017zero,yi2018neural}.

Other examples are instruction-following in the Neverwinter Nights game \cite{fleischman2005intentional}, dialogue about soccer videogames \cite{pasunuru2018game}, placing blocks appropriately given a final plan \cite{wang2016learning} and a more open ended building task using a grid of voxels \cite{wang2017naturalizing}. %
In the latter two cases the communication is one-sided with only the human issuing instructions, rather than dialogue, with the agent only able to act. %

There are also setups that consider static language and perception, for example 
image captioning \cite{lin2014microsoft},
video captioning \cite{yu2016video}, visual QA \cite{antol2015vqa} and
visual dialogue \cite{das2017visual,shuster2018engaging,mostafazadeh2017image}.
While grounded, the agent has no ability to act in these tasks. Talk the Walk ~\cite{devries2018talk} introduces a navigation game that involves action, perception and two-way dialogue, but is limited to small grids.

In summary, compared to many setups, our framework allows learning from both actions and (two-way) dialogue,
while many existing simulations typically address one or the other but  not both. 
In addition, being based on a gaming setup, our hope is that LIGHT can be fun for humans to interact with, enabling future engagement with our models. {All utterances in LIGHT are produced by human annotators, thus inheriting properties of natural language such as ambiguity and coreference, making it a challenging platform for grounded learning of language and actions.}

\section{LIGHT Environment and Task Setup}
\label{sec:light-env}

LIGHT is a large-scale, configurable text adventure environment for research
on learning grounded language and actions. 
It features both humans and models as embodied agents within a multi-player
fantasy MUD (multi-user dungeon)-like \cite{dieterle2009multi} environment.

To facilitate natural human-sourced 
(fantasy)
situations described by natural language, 
almost the entire environment is crowdsourced, including
locations, objects and their affordances, characters and their
personalities, and most importantly character interactions: dialogues and actions.  These components are collected through a series of annotation tasks that we will now describe.
These tasks are designed so that they can be combinatorially recombined. Data quality was maintained by requiring annotators to take a test (see Appendix \ref{appendix:data-quality}).
Overall statistics of the collected elements are given in Table \ref{table:data_stats}.
This environment can then be used to both train agents, and to evaluate 
them \textit{in situ} via their online interactions. 


\begin{table}[t]
\begin{center}
\small
\begin{tabular}{l|cccc}
 \toprule
Split                &       &       & Test &  Test\\
                     & Train & Valid & Seen & Unseen\\
\midrule                
Locations             & 589   & 352  & 499   & 74 \\
Objects              & 2658  & 1412   & 1895 & 844  \\
Characters           & 1369  & 546   & 820 & 360 \\
\midrule
Dialogues            & 8538 & 500 & 1000 & 739 \\
Utterances           & 110877 & 6623 & 13272 & 9853   \\
Emotes               & 17609 & 1156 & 2495  & 1301  \\
Actions              & 20256 & 1518 &  3227 & 1880 \\
\midrule
Vocabulary Size        & 32182 & 11327 & 11984 & 9984\\
Utterance Length   & 18.3 & 19.2  & 19.4 & 16.2  \\
\bottomrule
\end{tabular}
\caption{LIGHT dataset statistics.
\label{table:data_stats}
}
\end{center}
\end{table}

\begin{table*}[t]
  \begin{center}
    \begin{footnotesize}
        \begin{subtable}{\textwidth}
        \centering
        \captionsetup{width=.8\linewidth}
          \begin{tabular}{l|l}
            \toprule
            \textbf{Category:} & Graveyard\\
            \midrule
            \textbf{Description:} & Two-and-a-half walls of the finest, whitest stone stand here, weathered by the passing of \\
            & countless seasons. There is no roof, nor sign that there ever was one. All indications are \\
            & that the work was abruptly abandoned. There is no door, nor markings on the walls. Nor \\
            & is there any indication that any coffin has ever lain here... yet.\\
            \midrule
            \textbf{Backstory:} & Bright white stone was all the fad for funerary architecture, once upon a time. It's difficult \\
            & to understand why someone would abandon such a large and expensive undertaking. If they \\
            & didn't have the money to finish it, they could have sold the stone, surely - or the mausoleum \\
            & itself. Maybe they just haven't needed it yet? A bit odd, though, given how old it is. Maybe \\
            & the gravedigger remembers... if he's sober.\\
            \midrule
            \textbf{Neighbors:} & Dead Tree, south, following a dirt trail behind the mausoleum\\
             & Fresh Grave, west, walking carefully between fallen headstones\\
            \midrule
            \textbf{Characters:} & gravedigger, \textit{thief, peasant, mouse, bat}\\
            \midrule
            \textbf{Objects:} & wall, \textit{carving, leaf, dirt}\\
            \bottomrule
          \end{tabular}
          \subcaption{Example room created from the room collection and labelling tasks. Labels in italics were noted by workers as possibly present but not explicitly listed in the description or backstory.}
          \label{table:world-room}
          \end{subtable}
          \newline
          \vspace{2mm}
          \newline
          \begin{subtable}{\textwidth}
          \centering
          \resizebox{0.6\textheight}{!}{
          \begin{tabular}{l|l|l}
            \toprule
            \textbf{Character:} & Thief & Gravedigger\\
            \midrule
            \textbf{Persona:} & I live alone in a tent in the woods. & I am low paid labor in this town. \\
            & I steal food from the townspeople and & I do a job that many people shun because of \\
            & coal from the blacksmith. & my contact with death. \\
            & The village police can not find me & I am very lonely and wish I had someone\\
            & to put me in jail. &  to talk to who isn't dead.\\
            \midrule
            \textbf{Description:} & The thief is a sneaky fellow who takes from the & You might want to talk to the gravedigger, specially \\
            & people and does so in a way that disturbs the & if your looking for a friend, he might be odd but you\\
            & livelihood of the others. & will find a friend in him.\\
            \midrule
            \textbf{Carrying:} & meat, potatoes, coal & shovel\\
            \midrule
            \textbf{Wearing:} & dark tunic, cloak & \textit{nothing annotated}\\
            \midrule
            \textbf{Wielding:} & knife & \textit{nothing annotated}\\
            \bottomrule
          \end{tabular}
          }
          \subcaption{Example characters annotated via character collection tasks.}
          \label{table:world-characters}
          \end{subtable}  
          \newline
          \vspace{2mm}
          \newline
          \begin{subtable}{\textwidth}
          \centering
          \begin{tabular}{l|l|l}
            \toprule
            \textbf{Object} & \textbf{Description} & \textbf{Tags}\\
            \midrule
            shovel & The shovel is made of metal and silver. It is quite sturdy and appears new. & gettable, wieldable \\
            \midrule
            wall & The wall is pure white, the richest of which you have ever seen. & \textit{none} \\
            \bottomrule
          \end{tabular}
          \subcaption{Example objects annotated via object collection tasks}
          \label{table:world-objects}
          \end{subtable} 
          \caption{Example entities from the {\sc LIGHT} environment. Each was collected via tasks described in Section \ref{sec:light-env}.
     \label{table:world}}
    \end{footnotesize}
  \end{center}
\end{table*}

\paragraph{Locations}

We first crowdsourced a set of \numlocations game location settings from a base set of \numcategories categories ({\em countryside}, {\em forest}, {\em inside/outside castle}, {\em shore}, {\em graveyard}, {\em bazaar}, \dots -- full list in Appendix \ref{appendix-sec:location-categories}) which were selected by us to provide both inspiration and cohesion to annotators. 
Workers were provided a category and asked to create a description, backstory, names of connected locations, 
and contained objects and characters. See Table \ref{table:world-room} for an example.
Many descriptions are quite detailed, and there are clear semantics between entities (e.g. alligators being in swamps, cacti in a desert).

As all remaining tasks build upon the locations created in this first step, we selected 6 location categories ({\em underwater aquapolis}, {\em frozen tundra}, {\em supernatural}, {\em magical realm}, {\em city in the clouds}, and {\em netherworld}) designed to be distinct from the others to provide an isolated set of locations, characters, and objects for testing. These will be used to build what we  refer to as an {\em unseen} test set. 

Each location is collected independently, with the eventual aim that they can be glued together as desired to randomize world generation.
In this work, we consider actions and dialogues within a single location, so building a world map is not necessary. However, we will show that the environment has considerable influence on the dialogue, actions and 
grounded learning of models.

\paragraph{Characters}

We crowdsourced 1755 game characters from animals to trolls and orcs to humans of various types (wizards, knights, village clerk). 
See Table \ref{table:world-characters} for detailed examples.
Each character has a textual description, a persona (defined as a set of 3-5 profile sentences describing their traits, modeled after the Persona-Chat dataset \cite{zhang2018personalizing}),
and a set of objects that are currently being carried, wielded, or worn.
We sourced this list of characters to annotate from the ones provided in the location creation task.

\paragraph{Objects}

We crowdsourced 3462 objects, 
each with a textual description, and a set of affordances (whether it is a container, can be picked up, has a surface, is a weapon, is wearable, is food, is a drink).
See Table \ref{table:world-objects} for examples.
As before, we sourced this list of objects to annotate from the ones annotated for the locations and characters.

\paragraph{Actions and Emotes}
There are a set of actions in the game consisting of physical manipulations, and a set of emotes that display feelings to other characters, in line with existing MUDs.

Physical actions include
 {\em get}, {\em drop}, {\em put}, 
{\em give}, {\em steal}, 
{\em wear}, {\em remove}, {\em eat}, {\em drink}, {\em hug}
and {\em hit}, each taking either one or two arguments, e.g. {\em put robes in closet}. Every action has an explicit unambiguous effect on the underlying game state, and can only be executed if constraints are met, e.g. if the agent is holding the robes in the latter example.

Emotes include {\em applaud, blush, cringe, cry, dance, frown \dots, sulk, wave, wink} (22 in total) 
and have no effect on the game state other than to notify nearby characters of the emote, which can have effects on their behavior.
See Appendix \ref{appendix:detailed-actions-emotes} for further detailed descriptions.

\paragraph{Interaction}

Now that we have a fully realized underlying environment, we can attempt to learn and evaluate agents that can act and speak within it.
For this, we collect a human-human dataset of episodic interactions within the environment. 

For each dialogue, we place two characters in a random location (either two characters that were already assigned to it, or else randomly assigned characters), complete with the objects assigned to the location and to those characters.
Each character has access to their persona, the location description, and the objects present, and the interaction episode begins. The two characters take turns within the episode, and can execute one action (physical action or emote) and produce one dialogue utterance on each turn.
We crowdsourced 10,777 dialogues. Examples are given in Figure \ref{figure:dialogues} and Appendix Figures \ref{appendix:dialogues-first}-\ref{appendix:dialogues-last}.

\paragraph{Seen and Unseen Test Sets}
We provide two distinct test sets. The \textit{seen} test set consists of 
dialogues set in the
same world (set of locations) as the training set,
thus also consists of characters, objects, and personas that can appear in the training data. In contrast, the \textit{unseen} test set is comprised of dialogues collected on the unseen set of locations.
The unseen test set allows for evaluation of generalization capability to unseen topics in a similar domain and as we shall see, 
provides a more challenging test for current techniques.

\begin{figure*}[t]
  \begin{center}
    \begin{footnotesize}
      \begin{tabular}{l|l}
        \toprule
        \textbf{Persona 1: Servant} & \textbf{Persona 2: King}\\
        \midrule
        I come from the lower class.  & I am a king of the whole empire.\\
        I do what I am told without question. & I give rules and pursuit them. \\
        I can not read.  & I am brave and fearless. \\
        I have not seen my family in a long time. & \\
        \midrule
        Carrying: a duster, a small bucket, a rag & Carrying: a crown, a scepter\\
        Wearing: a shirt\\
        \midrule
        \multicolumn{2}{l}{\textbf{Setting:} The main foyer is massive. A grand staircase sits to the back of the foyer leading to the upstairs. }\\
        \multicolumn{2}{l}{At the front of the foyer stand two servants ready to help anyone who comes to visit. To the left of the }\\
        \multicolumn{2}{l}{room there is a doorway leading into a corridor. To the right there is a door leading to another corridor}\\
        \multicolumn{2}{l}{for the King's servants. At the foot of the stairs there is a bearskin rug that is staring at you almost as }\\
        \multicolumn{2}{l}{if still hungry. The walls are lined with portraits of the king and his family.}\\
        
        \bottomrule
        
        \multicolumn{2}{l}{\textbf{Servant:} my humble king. What am I to do to serve you? }\\
        \multicolumn{2}{l}{\textbf{King:} Ahhh. My loyal servant. Polish my scepter. }\\
        \multicolumn{2}{l}{\textit{~~Action:} give scepter to servant}\\
        \multicolumn{2}{l}{\textbf{Servant:} Yes my lord. I will polish it immediately. Am I to return it to you personally? }\\
        \multicolumn{2}{l}{\textit{~~Action:} put scepter in small bucket}\\
        \multicolumn{2}{l}{\textbf{King:} Yes. Yes. Of course. Also check the jewels in my crown. They seem loose. }\\
        \multicolumn{2}{l}{\textit{~~Action:} give crown to servant}\\
        \multicolumn{2}{l}{\textbf{Servant:} But sire I am not qualified to do that. Would you prefer I take it to someone? }\\
        \multicolumn{2}{l}{\textbf{King:} Oh fine then. }\\
        \multicolumn{2}{l}{\textit{~~Emote:} gesture sigh}\\
        \multicolumn{2}{l}{\textbf{Servant:}  I am sorry sir the rug startled me}\\
        \multicolumn{2}{l}{\textit{~~Action:} drop crown}\\
        \multicolumn{2}{l}{\textbf{King:} Haha! That's bear I slain on my latest hunting trip. He's a mighty beast!}\\
        \multicolumn{2}{l}{\textit{~~Emote:} gesture laugh}\\
        \multicolumn{2}{l}{\textbf{Servant:} and if I may ask where did you go hunting sire? }\\
        \multicolumn{2}{l}{\textbf{King:} The great woods of course. This bear was stealing children in the kingdom. Surely you heard about it.}\\
        \multicolumn{2}{l}{\textbf{Servant:} sire. I have not been outside of these walls in quiet some time. I have not seen my family in ages. }\\
        \multicolumn{2}{l}{\textbf{King:} Such is the life of a servant I suppose. How's that scepter looking?}\\
        \multicolumn{2}{l}{\textbf{Servant:} it is almost ready sire. and the crown who would you like me to take it to? }\\
        \multicolumn{2}{l}{\textit{~~Action:} get scepter from small bucket}\\
        \multicolumn{2}{l}{\textbf{King:} Here just give it back. I'll have the queen find someone.}\\
      \bottomrule 
      \end{tabular}
      \caption{Example dialogue from the {\sc LIGHT} dataset. 
 \label{figure:dialogues}}
    \end{footnotesize}
  \end{center}
\end{figure*}



\section{Learning Methods}

We consider a variety of models that can predict actions, emotes and dialogue, and explore the importance of grounding upon the location, objects, and other characters within the setting. For all models, we represent context as a large text sequence with a special token preceding each input type (persona, setting, self emote, partner emote, etc.). We work with two model classes: \textit{ranking} models that output the maximal scoring response from a set of potential candidate responses and \textit{generative} models that decode word by word. 

\paragraph{Baseline Ranking Methods} 
We report a Random baseline (selecting a random candidate from the candidates)
and an Information Retrieval (IR) baseline that uses word overlap with TF/IDF weighting. We use \textit{Starspace} \cite{wu2018starspace} which learns a bag-of-words embedding for context and candidates to maximize the inner product of the true label using a ranking loss. Lastly, we use \textit{fastText} \cite{joulin2016bag} to classify which emote should be predicted next as there are only 22 classes. Finally, we compare the performance of our best models to human performance on each of the prediction tasks.

\begin{table*}[t]
\begin{center}
\resizebox{\textwidth}{!}{
\begin{tabular}{l|lllllll}
\multicolumn{1}{l}{\bf Query:} &\textbf{chicken} &\textbf{pirate} &\textbf{coffin} &\textbf{rake} &\textbf{tavern} &\textbf{meadow} \\
\hline
\parbox[t]{2mm}{\multirow{6}{*}{\rotatebox[origin=c]{90}{objects}}} &chicken coop &Pirate swords &the remains &shovel &Ale bottles &flower pot \\
&eggs &dock &remains &garden &beer &fruit \\
&a pen for the chickens &cargo &bones &a garden &mug of mead &An enchanted amulet. \\
&chimney &ship &bones of the innocent &Hand carved stone &a large ornate table &citrus fruit \\
&corn &seagulls on the dock &adventurer's remains &garden bench &beer keg &fruit trees \\
&stone chimney &cargo crates &precious jewels &small garden &mug &nice fruit trees \\
\hline
\parbox[t]{2mm}{\multirow{6}{*}{\rotatebox[origin=c]{90}{characters}}} &chickens &boat captain &spirits of our ancestors &gardener &tavern owner &a deer \\
&fox trying to steal chickens &captain &mourner &stable hand &bartender &a songbird \\
&farmers &merchant &zombies &Garden dog &Goblin King's bartender &fruit bats \\
&The farmers &boat workers &families &stable boy &A serving wench &parent \\
&farmer &workers &bandit &A stable boy &Serving wench &butterfly \\
&poorer subsistence farmers &the workers &the royal family &two guards &a round man with a bushy mustache &Small insects \\
\hline
\parbox[t]{2mm}{\multirow{6}{*}{\rotatebox[origin=c]{90}{locations}}} &Chicken Pen &Pirate Ship &Old Crypt &Across the King's Garden &The werewolves tavern &Lush meadow \\
&Corn field &Dock at the Port &sacristy &Hidden garden &Tavern of Browntavia &Flower Field \\
&Farmer's house &Loading Dock &Disposal area &The garden courtyard &Port Tavern &flower garden \\
&Large Farm &Fishing Dock &inside temple crypt &Church garden &The bar &Mushroom Hut \\
&Pig Pen &crew berthing &Sacrifice Chamber &Tool Shed &bazaar outside the royal city &Archery zone \\
&The old red barn &captain's cabin &Shrine of Sretniy &flower garden &Outside gates &The witches cottage \\
\hline
\parbox[t]{2mm}{\multirow{6}{*}{\rotatebox[origin=c]{90}{actions}}} &get chicken &hug pirate &put torch in coffin &get rake &hug tavern owner &get flower from meadow \\
&hug chicken &hit pirate &get torch from coffin &drop Rake &give food item to tavern owner &put flower in Meadow \\
&hit chicken &steal sword from pirate &put bone in coffin &steal Rake from gardener &give telescope to tavern owner &give Flower to a deer \\
&give cowbell to chicken &steal cargo from pirate &get bone from coffin &give Rake to thing &drink drink &give Flower to deer \\
&steal sword from chicken &give cargo to pirate &hit archaeologist &give Rake to person &drop drink &steal Flower from a deer \\
&give corn to chicken &give Daggers to pirate &hug archaeologist &give Rake to guard &give coin to bartender &get flower \\
\hline
\parbox[t]{2mm}{\multirow{6}{*}{\rotatebox[origin=c]{90}{vocabulary}}} &bock &crew &archaeologist &vegetable &drink &flower \\
&tasty &ye &robber &carved &drinks &amulet \\
&bawk &port &crypt &alice &regular &songbird \\
&moo &sea &loss &hook &item &wasp \\
&egg &seas &adventures &exorcisms &tip &an \\
&lay &sail &earn &tomatoes &bottles &holiness \\
\hline
\end{tabular}
}
\caption{
Neighboring Starspace phrase embeddings (no pretraining from other data) for different types of entities and actions. The first row are arbitrarily chosen queries  (chicken, pirate,  coffin, rake, tavern, meadow), and the subsequent rows are their nearest objects, agents, locations, actions and vocabulary in embedding space.
\label{starspaceeee}
}
\end{center}
\end{table*}

\paragraph{Transformer Memory Network} We use the transformer memory-based ranking model from 
\citet{dinan2019wizard}. It uses a transformer \cite{vaswani2017attention}  to produce separate representations (memory slots) for each sentence from the grounding information (setting, persona, objects). It then performs attention given the dialogue context over the memories to produce a context embedding, which is used to score candidates via the dot product with the transformer-based representation of the candidate.
At training time, other samples in the batch are used as negative candidates. 
For emote prediction, we train by ranking against the full set of possible emotes as there are only 22 distinct classes.

\paragraph{BERT Bi-Ranker and Cross-Ranker} We adapt the BERT pretrained language model \citep{devlin2018bert} to the tasks of dialogue and action prediction.
 We explore two architectures for leveraging BERT. First, we use the \textit{BERT-based Bi-Ranker} to produce a vector representation for the context and a separate representation for each candidate utterance. This representation is obtained by passing the first output of BERT's 12 layers through an additional linear layer, resulting in an embedding of dimension 768.
 It then scores candidates via the dot product between these embeddings and is trained using a ranking loss. 

Second, the \textit{BERT-based Cross-Ranker} instead concatenates the context with each candidate utterance, similar to \citet{wolf2019transfertransfo}. Then, each candidate is scored by computing a softmax over all candidates. 
Unlike the BERT-based Bi-Ranker, the concatenation of the context with each individual candidate allows the model to attend to the context when encoding each candidate, building a \textit{context-dependent} representation of each candidate. In contrast, the Bi-Ranker can use self-attention to build the candidate and context representations, but cannot modify their representation based upon the context. However,  the Cross-Encoder is far more computationally expensive 
($\sim$11,000 slower than the Bi-Ranker for dialogue retrieval)
as each concatenated representation must be recomputed, while the Bi-Ranker can cache the candidates for reuse (see Appendix \ref{appendix:bert-speed}).

\paragraph{Generative Models} Similarly to the ranking setting, we use the Transformer Memory Network from \citet{dinan2019wizard} to encode  the context features (such as dialogue, persona, and setting). However, to predict an action, emote, or dialogue sequence, we use a Transformer architecture to decode while attending to the encoder output. 

For the task of action generation, the set of candidates for ranking models to rank the true action sequence against is constrained by the set of valid actions. For example, the character cannot \textit{pick up book} if there is no book.  In the generative model, we compute the log likelihood for the set of possible candidates and normalize to constrain the output space to valid actions to improve the results.

\subsection{Implementation}

We implement models using PyTorch in ParlAI \cite{miller2017parlai}. 
Ranking Transformer models are pretrained on Reddit data \cite{training_millions} and fine-tuned. We use the BERT \cite{devlin2018bert} 
implementation  provided by Hugging Face\footnote{\scriptsize https://github.com/huggingface/pytorch-pretrained-BERT}
with pre-trained weights, then adapted to our Bi-Ranker and Cross-Ranker setups.
 Generative models are pretrained on the Toronto Books Corpus and fine-tuned except for emote prediction which does not leverage pretraining. We apply byte-pair encoding \cite{sennrich2016neural} to reduce the vocabulary size for generative models. 
We decode using beam search with beam size 5. 

\begin{table*}[t!]
\begin{center}
\small
\begin{tabular}{l|rrr|rrr}
 \toprule
                     & \multicolumn{3}{c}{Test Seen} & \multicolumn{3}{|c}{Test Unseen} \\
                     & Dialogue & Action & Emote & Dialogue & Action & Emote \\
 Method              & R@1/20 & Acc    & Acc & R@1/20 & Acc    & Acc\\
\midrule                
Random baseline   & 5.0 & 12.2 & 4.5 & 5.0 & 12.1 & 4.5\\
IR baseline       & 23.7 & 20.6& 7.5  & 21.8 & 20.5 & 8.46 \\
Starspace         & 53.8 & 17.8 & 11.6 & 27.9 & 16.4 & 9.8  \\
Transformer MemNet & 70.9	& 24.5 &	17.3  & 66.0 & 21.1 & 16.6 \\
BERT-based Bi-Ranker & \textbf{76.5} & 42.5 & 25.0& 70.5 & 38.8 & 25.7 \\
BERT-based Cross-Ranker & 74.9 & \textbf{50.7} & \textbf{25.8}& 69.7  & 51.8 &28.6 \\
\midrule 
\rule{0pt}{2ex}Human Performance* &  *87.5 & *62.0 & *27.0 &  *91.8 & *71.9 & *34.4  \\
\bottomrule
\end{tabular}
\caption{Ranking model test performance.
(*) Human performance is computed on a subset of data.
\label{table:main_resulteroosas}
}
\end{center}
\end{table*}

\begin{table}[t!]
\resizebox{\columnwidth}{!}{
\begin{tabular}{l|cccc}
 \toprule
                     & Dialogue & Action & Emote\\
                     & R@1/20 & Acc    & Acc \\
\midrule                
BERT-based Bi-Ranker        & 76.0  & 38.7   & 25.1 \\
~~~~actions+emotes only     & 58.6  & 18.3   & 10.6  \\
~~~~dialogue only           & 68.1  & 39.4   &  23.6 \\
~~~~dialogue+action+emote   & 73.2 & 40.7  & 23.1 \\
~~~~dialogue+persona       & 73.3  & 41.0 & 26.5 \\
~~~~dialogue+setting        & 70.6  & 41.2 & 26.0 \\
~~~~dialogue+objects        & 68.2  & 37.5 & 25.5 \\
\bottomrule
\end{tabular}
}
\caption{BERT-based Bi-Ranker ablations (valid set). The LIGHT environment includes a variety of grounding information: dialogue, action, emote, persona, setting, and object descriptions.
\label{table:ranking_ablations}
}
\end{table}

\begin{table}[t]
\resizebox{\columnwidth}{!}{
\begin{tabular}{l|cccc}
 \toprule
                     & \multicolumn{2}{c}{Dialogue}   & Action & Emote\\
                     & PPL & F1 & Acc & Acc \\  
\midrule                
Generative Transformer  & \textbf{27.1} & \textbf{13.9} & \textbf{13.0} & 20.6 \\
~~~actions+emotes only  & 32.8 & 9.3 &  10.5 & 15.3 \\
~~~dialogue only        & 28.0 & 12.5 & 12.3 & 20.0 \\
~~~dialogue+action+emote& 27.6 & 12.3 & 12.8 & \textbf{22.0} \\
~~~dialogue+persona    & 27.8 & 12.9 & 12.3 & 20.8 \\
~~~dialogue+setting     & 27.8 & 12.1 & 11.5 & 17.8 \\
~~~dialogue+objects     & 27.7 & 12.8 & 11.0 & 20.2 \\
\bottomrule
\end{tabular}
}
\caption{Generative Transformer ablations (valid set).
\label{table:generative_ablations}
}
\end{table}

\subsection{Evaluation}

\paragraph{Automatic} To evaluate our models, we calculate percentage accuracy for action and emote prediction. 
For dialogue, we report Recall@1/20 for ranking the ground truth among 19 other randomly chosen candidates for ranking models and perplexity and unigram F1 for generative models.

\paragraph{Human} We present humans with the same ranking task and report R@1/20 to estimate their performance on this task. During the evaluation, we provide annotated examples on the training in addition to  examples on the test set. We only keep the annotations of evaluators who had high accuracy on the training examples to filter low-accuracy evaluators. The training accuracy bar was selected due to the difficulty of the separate tasks. 
Our methods for human evaluation are described in more detail in Appendix \ref{appendix:human-eval-stats} along with how many turns were evaluated.

\section{Results}

The ranking models are compared in Table \ref{table:main_resulteroosas} on the seen and unseen test sets,  and ablations are shown for both the BERT-based Bi-Ranker and Generative Transformer in Tables \ref{table:ranking_ablations} and \ref{table:generative_ablations}.

\subsection{Comparison of Models and Baselines}

The IR baseline shows non-random performance, but is outperformed by Starspace  which is a stronger baseline. We also tried {\textit{FastText}} on the emotion task which gave a seen test accuracy of 13.2.
Transformer architectures prove significantly stronger at all tasks, with BERT pretraining proving important for best results as used in the Bi-Ranker and Cross-Ranker architectures. The latter, which can create a context dependent representation of each label candidate,  
is better at actions and emotes. Human performance is still above all these models, leaving space for future improvements in these tasks.
The generative  Transformer model did not work as well using these metrics. 

\subsection{Generalization Capability on Unseen Test}

The six new unseen test settings  are a slightly easier task in absolute numbers (Table \ref{table:main_resulteroosas}, right), with improved scores for humans and some models. 
We observe that BERT-based models exhibit good transfer ability relative to other models, but the gap between their performance and human performance increases from the seen test set to the unseen one. Specifically, there is a 21 point gap on the unseen dialogue test set compared to an 11 point gap on the seen test set, making
this a significant challenge for future methods.

\begin{table*}[t]
  \begin{center}
    \begin{footnotesize}
      \begin{tabular}{l|l}
        \toprule
        \multicolumn{2}{l}{\textbf{Persona:}  I am a part of a group of travelers.  I go from town to town selling food to the locals.  }\\
        \multicolumn{2}{l}{I grew up poor, but my travels have paid off well.}\\
        \midrule 
        \textbf{Setting 1:  Fishmonger's stall, Port} & \textbf{Setting 2: Dunes, Desert}\\
        A small booth near the edge of the port, it's protected by a & A massive hilly landscape that is nothing but sand and a few \\
        piece of old, sun-bleached sailcloth. Baskets of freshly- & rocks.  As you walk this area, you can find some human and \\
        caught fish, bivalves, and eels sit in the shade in stained  & animal remains along with broken down wood wagons.\\
        wooden troughs of water. A small, aggressive-looking dog & \\
        is chained to one table, presumably to keep cats away. & \\
        The stall is redolent with the aroma of fish. & \\
        \bottomrule
        \textbf{Friend:} I wonder what I could eat around here... & \textbf{Friend:} I wonder what I could eat around here...  \\
        \textit{~~Emote:} ponder & \textit{~~Emote:} ponder \\
        \textbf{Traveler:} Customer, are you here shopping for fish too? & \textbf{Traveler:} Well, the desert is certainly the wrong place \\ 
        & for you my friend. \\
        \textbf{Friend:} What brings you to this place? & \textbf{Friend:} What brings you to this place? \\
        \textbf{Traveler:} I like to come around here for food. Sometimes & \textbf{Traveler:} I am travelling to the castle market to sell my goods. \\ 
        people who travel through drop the most delicious things. &  I have a terrible sense of direction and have been wondering in \\
        Once in a while it's roasted meet or fish. & the sweltering heat for hours until I found your Oasis. \\
      \bottomrule 
      \end{tabular}
      \caption{Predicted dialogue by the BERT-based Bi-Ranker (as the \textit{traveler} character) given different settings.
 \label{table:setting-dialogues}}
    \end{footnotesize}
  \end{center}
\end{table*}

\begin{table}[t!]
\begin{small}
\begin{center}
\begin{tabular}{l | l l}
\toprule
\multicolumn{3}{l}{Self name: Sea Witch.}\\
\multicolumn{3}{l}{Self Previous Dialogue: What do you know about that} \\
\multicolumn{3}{l}{knight standing over there?} \\
 \midrule
{\bf Input Dialogue + Emote} & \textbf{Partner}           &{\bf Prediction}\\

His armor is garrish. You  & Mermaid & laugh \\
know I don't fraternize   & Thief & frown \\
 with land dwellers, \textit{pout} & & \\
\hline
He is a terrible knight & Mermaid & scream \\
and I hate him, \textit{cry} & Troll & laugh \\
\hline 
I will battle him until the  & Mermaid & stare\\
end of my days, \textit{scream} & Orc & nod \\
\bottomrule
\end{tabular}
\caption{
Predicted emotes by the Generative Transformer given example inputs from dialogue partner. 
\label{table:emote_examples}
}
\end{center}
\end{small}
\end{table}

\subsection{Data Inter-connectedness and Coverage}

To illustrate the coverage of entities and actions in the LIGHT world, and the inter-connectedness between them learnable from our data, we trained a simple Starspace embedding model with no prebuilt embeddings (so, on our data alone, thus precluding BERT) on all three tasks and show embeddings in Table \ref{starspaceeee}.
There is clearly a vast variety of learnable concepts and rich structure between characters, locations, objects, actions and the language describing them.
We also show additional t-SNE plots and heatmaps showcasing these relationships in Appendix \ref{app:tsne-etc}.

\subsection{Importance of Grounding}

\paragraph{Effect of Various Environment Features} We provide a large quantity of information about the environment to each of our models --- not only dialogue, but the description of the setting, the character's persona, present objects with descriptions, and more. We analyze the usefulness of the additional grounding information in Tables~\ref{table:ranking_ablations} and \ref{table:generative_ablations}.

For the dialogue task, having access to all of the environmental information provides the best performance for both retrieval and generative models. Training on dialogue alone substantially decreases performance, while each experiment that adds additional grounding information such as the past actions, persona or the setting description, improves the score. Providing object descriptions as a feature leads to the least improvement. As there are both a large quantity of objects that can be present and objects tend to have long descriptions, it can be challenging for the model to associate such information to a dialogue, action, or emote prediction task. 
The persona features were found to be impactful,
 which makes sense as they shape the things the character 
 says (and does).

Action sequence and emote prediction are much improved when using the dialogue history compared to using only past action history. 
Other features generally have lesser impact in this case, but still give some improvements. Including all features appears challenging for the model, perhaps because of the large input to attend over, resulting in improved results for some ablations.

Most importantly, for all tasks
training on the available dialogue data is necessary for good performance. Providing only the action and emote as context results in the worst performance, even on action and emote prediction tasks. Moreover, using dialogue and actions simultaneously improves results almost everywhere. The integrated environment in which agents can both act and speak to other agents provides relevant information that can be used across all tasks.



\begin{table}[t]
\begin{center}
\resizebox{0.5\textwidth}{!}{
\small
\begin{tabular}{ll}
\toprule
{\bf Input from Partner: Wizard}           &{\bf Prediction  {\scriptsize (Self name: Servant)}}\\
I'm feeling sad                    & hug wizard\\
You must die!                      & hit master wizard\\
Try putting on something else   & remove patterned outfit\\
I'd like you to feed me            & give food to master wizard\\
Can you grab me a paper            & give book to wizard's assistant\\
Can you grab me a beer             & get beer\\
Clean up                           & get duster\\
Hide the gold                    & put gold in satchel \\
\midrule
{\bf Input from different agents}     &  {\bf Prediction} \\
Wizard: Can I have some drink?   &  drop potion \\
Servant: Can I have some drink?  & give wine to servant \\
Bear: Can I have some drink?   & give water to bear \\
\bottomrule
\end{tabular}
}
\caption{
Predicted actions by the BERT-based Bi-Ranker given example inputs from the dialogue partner.
\label{table:action_examples}
}
\end{center}
\end{table}

\paragraph{Context affects predicted utterances}

We investigate the effect of the environmental context 
on the predictions by modifying the context and examining the changes in predicted dialogue, action, and emotes using the BERT-based Bi-Ranker. 

The input dialogue and speaker has a strong effect on the predicted action, as shown in Table~\ref{table:action_examples}, ranking over all training set actions. 
For example, the partner asking for an item results in a predicted action to retrieve it,
despite our dataset not being explicitly instructional, and is dependent on  who asks.

A similar effect is observed for emote prediction. Modifying the dialogue
and emote input produces a variety of different predicted emotes in Table~\ref{table:emote_examples}. Further, keeping the context otherwise fixed but modifying the partner name from \textit{mermaid} to \textit{orc} results in a different predicted emote --- the mermaid stating \textit{I will battle him} leads to a \textit{stare} while the orc receives a \textit{nod}. 

Finally, for dialogue prediction we find the model produces different outputs that are more appropriate for a given setting, even if the dialogue and characters 
are the same, see Table~\ref{table:setting-dialogues}. 
With the same text about food, the model retrieved dialogue that was setting appropriate. In the fishmonger's stall, it asked if the human agent was a customer shopping for fish, but in the desert dunes it suggested we might be looking in the wrong place.


\section{Conclusion}

We introduced a large-scale crowdsourced fantasy text adventure game research platform where agents---both models and humans---can act and speak in a rich and diverse environment of locations, objects, and other characters.
We analyzed a variety of models and their ability to leverage the grounding information present in the environment. We hope that this work  can enable future research in grounded language learning and further the ability of agents to model a holistic world, complete with other agents within it. 

\section{Acknowledgements}

We thank Ta\'is Mauk and Lisa Wong for their help with this project.


\bibliography{acl2019}
\bibliographystyle{acl_natbib}

\appendix
\section*{\LARGE Supplementary Material}
\vspace{1em}

\section{Model Inputs}
For extra clarity, we show here the exact input representation given to our models when including all the grounding features we consider in the experiments (setting, objects, characters + personas, actions, emotes, and dialogue). An example is given in Figure \ref{app:input_format}.

We note that there are other ways to represent this information that we have not explored that could improve performance. Further, there is additional information in LIGHT that could possibly be encoded in the input text: for example, what characters are carrying,  and the affordances of objects. The latter, while
 not explicitly provided in the input does constrain the available actions, so it is still used by the model.
 Object affordances such as \textit{is gettable} are visible to models
via the action history, but more explicit inputs could potentially be useful,
and this could be explored in future work.

\section{Bi-Ranker and Cross-Ranker Speeds}
\label{appendix:bert-speed}

We give test time computation speeds for the BERT-based Bi-Ranker and Cross-Rankers
in Tables \ref{bert-speed1} and \ref{bert-speed2} for the emote and dialogue tasks.
For the emote task, the Cross-Ranker is still feasible due to there being only
22 labels to compute, although it is still 4.6x slower than the Bi-Ranker if the 
22 candidate representations are cached. The Bi-Ranker can always cache label representations if they are fixed for many input examples (the common case) because the representation does not depend on the input. For the Cross-Ranker this cannot be done because the label representations are contextually dependent on the input.
For dialogue retrieval, because the number of candidates is so large (more than
100,000) caching makes the Bi-Ranker feasible whereas the Cross-Ranker, which cannot cache label representations, is infeasible to compute.

\begin{table}[h]
\begin{center}
\resizebox{\columnwidth}{!}{
\begin{tabular}{l|ll}
\toprule
 Emote           &     Bi-Ranker 	& Cross-Ranker \\
w/o caching     &   171s	    &  326s  ($\sim$1.9x slower)\\
with caching    &   70s         &  n/a   ($\sim$4.6x slower) \\
\end{tabular}
}
\end{center}
\caption{Bert Bi-Ranker and Cross-Ranker speeds on the emote task, test seen (2495 examples), 22 candidates per example.
\label{bert-speed1}
}
\begin{center}
\resizebox{\columnwidth}{!}{
\begin{tabular}{l|ll}
\toprule
            &     Bi-Ranker 	& Cross-Ranker \\
Dialogue    & 2.07s  &    24453s  ($\sim$11812x slower) \\
\end{tabular}
}
\end{center}
\caption{Bert Bi-Ranker and Cross-Ranker speeds on the dialogue task, per single example average (retrieval over  110,877 training set candidates).
\label{bert-speed2}
}
\end{table}

\begin{figure*}[t]
\begin{small}
\begin{tabular}{l}
{\bf Input to Model:}\\
\_task\_speech\\
\_setting\_name main foyer, Inside Castle\\
\_setting\_desc The main foyer is massive. A grand staircase sits to the back of the foyer leading to the upstairs.\\ 
At the front of the foyer stand two servants ready to help anyone who comes to visit. To the left of the room there\\
is a doorway leading into a corridor. To the right there is a door leading to another corridor for the King's servants.\\ 
At the foot of the stairs there is a bearskin rug that is staring at you almost as if still hungry. The walls are \\
lined with portraits of the king and his family.\\
\_partner\_name servant\\
\_self\_name king\\
\_self\_persona I am a king of the whole empire. I give rules and pursuit them. I am brave and fearless.\\
\_object\_desc a duster : The duster has large gray feathers bound together by a leather wrap.\\
\_object\_desc a small bucket : The bucket may be small but it gets the job done.\\
\_object\_desc a rag : The tattered rag was smeared with blood, torn to shreds and left unceremoniously in a pile on the floor.\\
\_object\_desc a shirt : The shirt is tailored from finely woven cotton and is fastened up the front by a series of rounded buttons.\\
\_object\_desc a crown : Thought of as a holy item, the crown goes only to those who are worthy enough.\\
\_object\_desc a scepter : On its handle, you see two red gems gleaming like eyes of an animal.\\
\_partner\_say my humble king. What am I to do to serve you? \\
\_self\_act give scepter to servant\\
\_partner\_say Yes my lord. I will polish it immediately. Am I to return it to you personally? \\
\_partner\_act put scepter in small bucket\\
\_self\_act give crown to servant \\
\midrule
{\bf Label:} Yes. Yes. Of course. Also check the jewels in my crown. They seem loose.\\
\end{tabular}
\end{small}
\caption{Example input format (and target label) given to models, following the same dialogue in Figure \ref{figure:dialogues}. Tokens like "\_setting\_name" are special tokens intended to be signifiers for 
the encoding module of a network to know which piece of grounding information is being read on that line. 
\label{app:input_format}
}
\end{figure*}

\section{Unseen Test Set Overlap}

The unseen test set is chosen by design to be relatively distinct from those available in the training set, and the actual content (descriptions, personas, dialogues) are entirely disjoint. However,
due to the large size of the dataset, it is possible the names of locations, characters, and objects in the unseen set could have word overlap. We assert this by comparing word overlap with the names of locations, characters, and objects in the training set. Of the 73 locations, 207 characters, and 956 objects created from the unseen location categories, the names of 3 locations, 96 characters, and 203 objects exactly match names of elements in the training set. We note that these represent names such as \textit{tavern}, but the chats are collected with the full location descriptions (which are unseen in the training set) and thus reduces overlap with train. 

\section{Crowdsourcing Methodology}
\label{appendix:crowdsource-details}
\label{appendix:data-quality}

Expanding on the dataset collection explanations in section \ref{sec:light-env}, a number of steps were taken to attain a level of quality and consistency. The first and most influential came from the constrains of the setting itself. We used a fantasy setting to try to encourage some kind of continuity across the dataset. We believed that workers would share some kind of common understanding about what a fantasy environment would entail, and then this understanding would be reflected in the dataset. It also ensured there were easy ways to flag certain workers that were creating content that wouldn't make sense in the dataset (referencing real locations, modern day objects, etc.). From here we could remove some content and filter workers out from continuing to work on this dataset. The other primary technique regarded using rounds of pilots and staged tasks to gradually filter towards high quality content rather than collecting all of the content in a single forward pass. Nearly half of the content in each initial pilot task was discarded, and we iterated on pilot tasks until the discard rate was less than 1 in 30 tasks. The rest of this section will discuss some specific measures taken at the individual task level, and will acknowledge some arguable deficiencies and potential areas of improvement on the dataset in its current form.

\paragraph{Locations}

\begin{figure*}
    \centering
    \includegraphics[width=\textwidth]{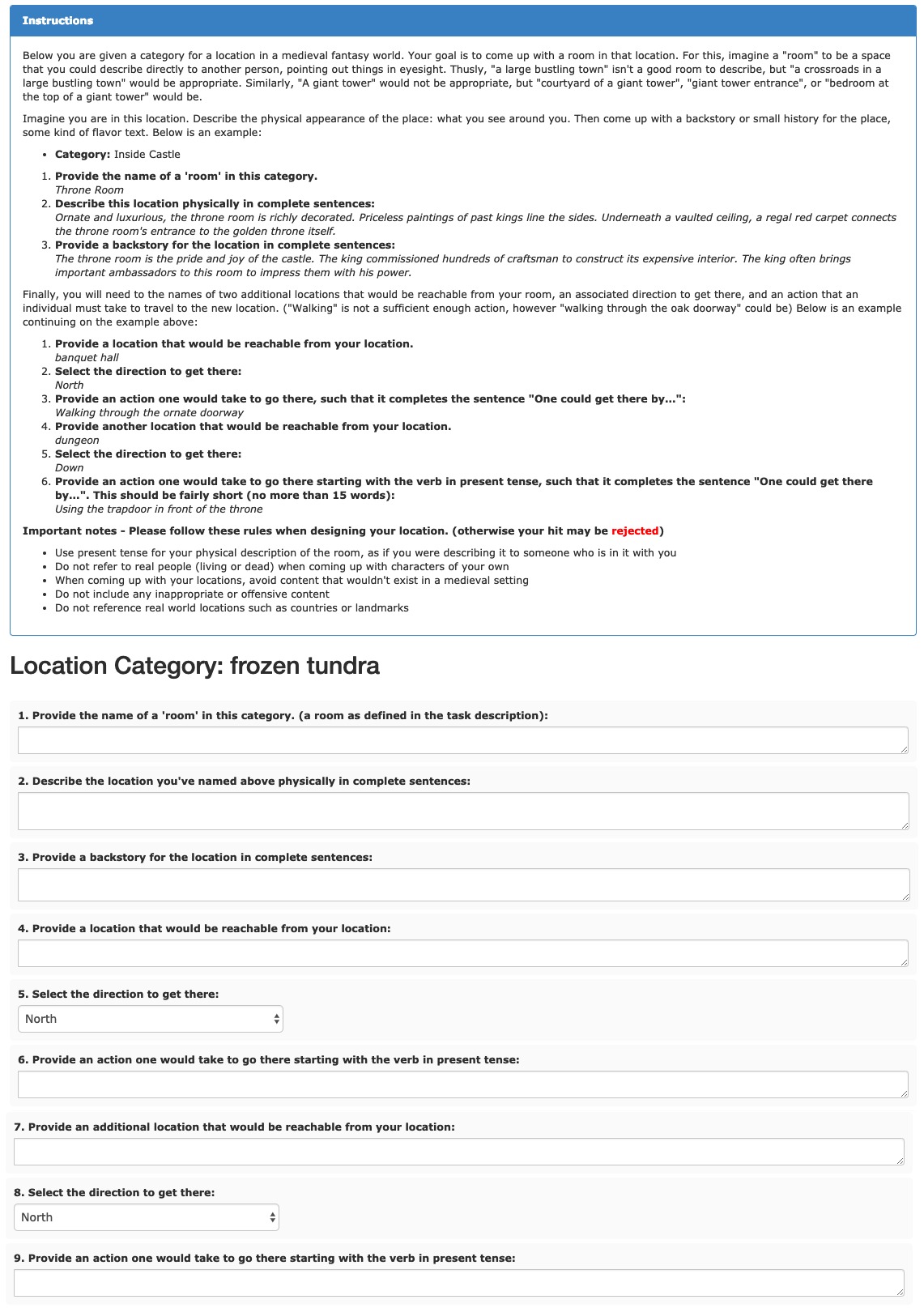}
    \caption{Form for Crowdsourcing Task 1}
    \label{fig:turk-task-1}
\end{figure*}

\begin{figure*}
    \centering
    \includegraphics[width=\textwidth]{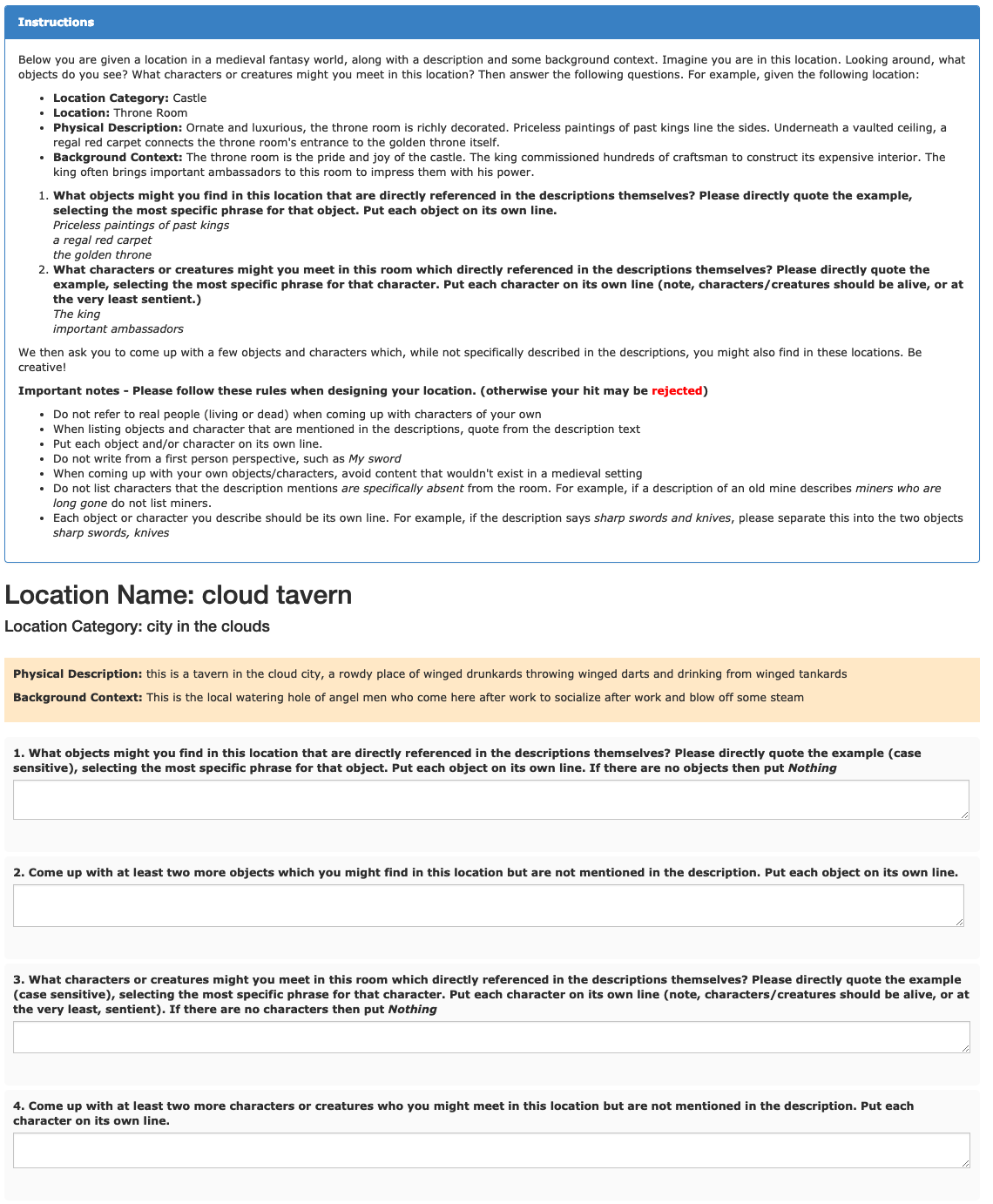}
    \caption{Form for Crowdsourcing Task 2}
    \label{fig:turk-task-2}
\end{figure*}

The location task of creating a description, backstory, list of connected rooms, and annotations of characters and objects present seemed to be too disjoint of a task based on the crowdsourcing best practice of breaking down tasks into as atomic of an action as possible. Thus we split it into two tasks, the first to provide the core text content and list of connected rooms, and the second to annotate the content inside those rooms. We will refer to these as \textit{Task 1} and \textit{Task 2}, and were simple form-entry tasks as displayed in Figures \ref{fig:turk-task-1} and \ref{fig:turk-task-2}. These two tasks were used in sequence to produce the locations present in the dataset.

In order to drive quality, we manually reviewed a handful of rooms from each worker to assert that the rooms had proper English descriptions and back-stories, and that the room fit appropriately in the category provided. In retrospect, given the two-tiered task setup and some of the techniques we developed later in the collection setup, we could have asked workers who were annotating rooms in Task 2 to provide some kind of signal about the quality of the rooms from Task 1 in order to have a lower-cost method for evaluating the quality of the work from Task 1 than using our own time.

Ultimately, one of the most important steps for improving dataset quality at this stage was creating form validators that caught the most common error cases from the first time around. These validators had the bonus effect of deterring botting of our tasks, as they couldn't pass the validation stage. For Task 1, the simple validator we ended up using asserted at least one complete sentence (determined via capitalization and punctuation) for both the description and background. For Task 2, our validation step forced workers to enter values that had direct word overlap with the entered text.

One of the largest difficulties with Task 2 was that some workers would optimize for grabbing key words out of the text without taking the time to fully understand the context. As thus, phrases like \textit{"and the remains of adventurers long dead"} would occasionally result in workers annotating the presence of \textit{adventurers} as characters in the given room. We attempted to mitigate this type of false positive with both explanatory examples and spot checks to soft-block workers who made this mistake consistently. At the moment a small number of these still remain in the dataset, but generally in instances where it still makes sense as in the above example, where the room definitely has remains of previous adventurers, but appropriately could also have some current adventurers as well.

\paragraph{Characters}

\begin{figure*}
    \centering
    \includegraphics[width=\textwidth]{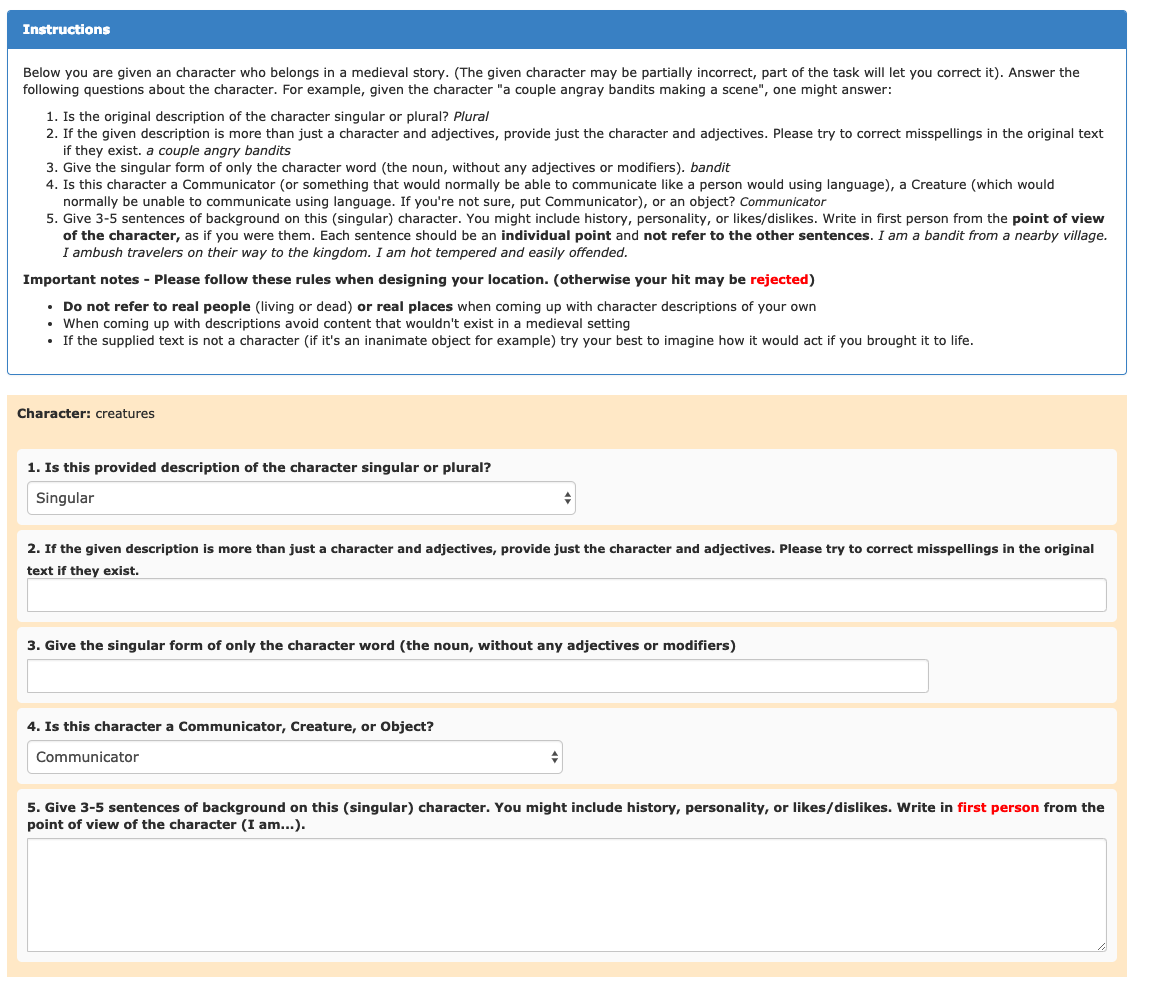}
    \caption{Form for Crowdsourcing Task 3}
    \label{fig:turk-task-3}
\end{figure*}

\begin{figure*}
    \centering
    \includegraphics[width=\textwidth]{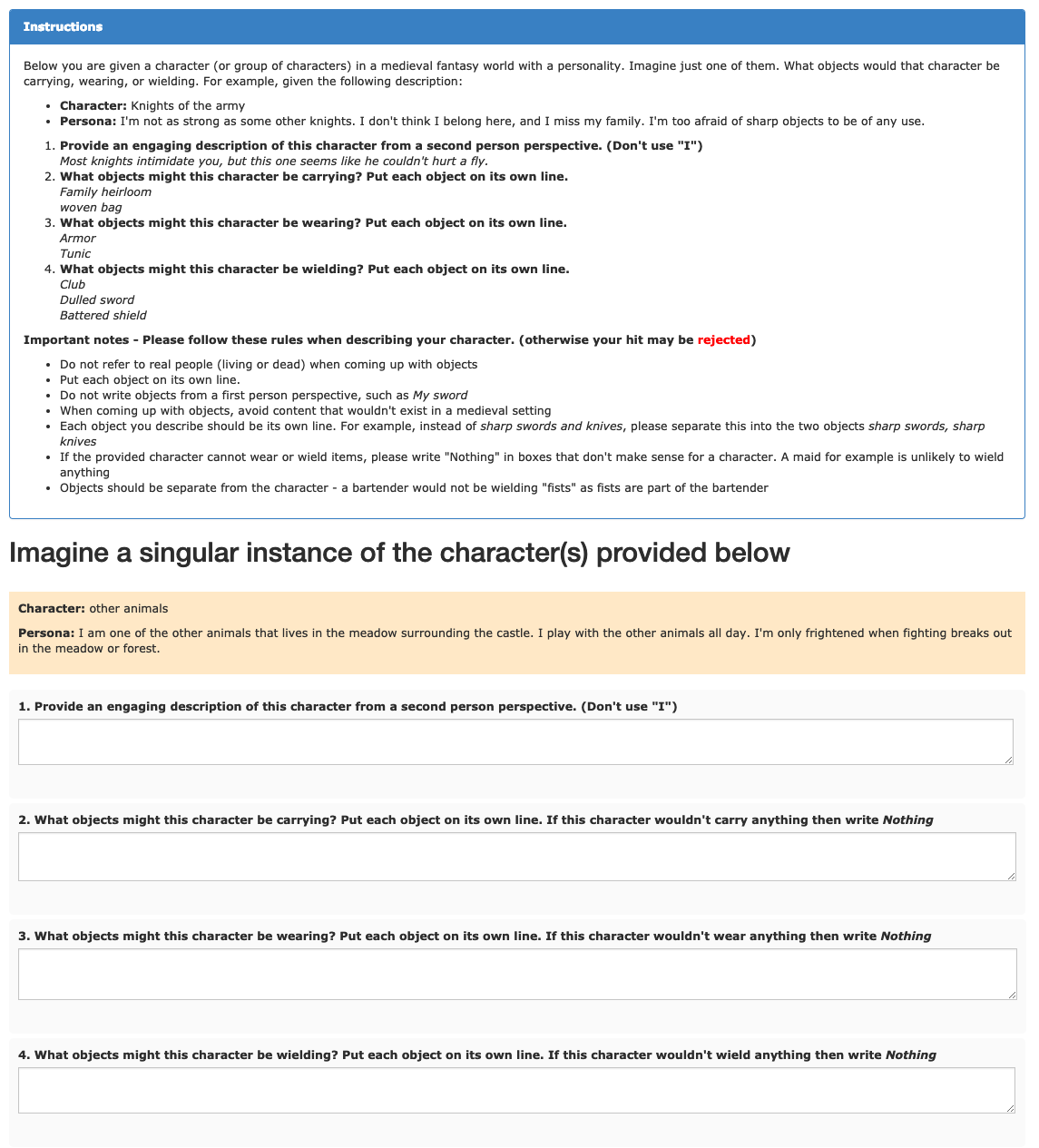}
    \caption{Form for Crowdsourcing Task 4}
    \label{fig:turk-task-4}
\end{figure*}

Similarly to how we split Location collection into two tasks, Character collection was split into two tasks as well. The first asked workers to clean up the span selected in Task 2 in order to remove words that didn't directly relate to or describe the character, and to provide a singular form for plural characters (as we intended for someone to eventually play the role of the singular character), tag the character as a person, creature, or object that was accidentally tagged as a character, and then asked for a \textit{first-person} perspective persona for the singular character. The second task gave workers the name of a character and their persona, and asked for a \textit{second-person} perspective description for the character as well as a list of objects that the character may be carrying, wielding, or wearing. We'll call these tasks \textit{Task 3} and \textit{Task 4}, and these were also collected via form-based tasks as displayed in Figures \ref{fig:turk-task-3} and \ref{fig:turk-task-4}. We used complete sentence form validation for both the persona from Task 3 and text descriptions in Task 4 to flag potential bad examples to filter out.

The goal of the Task 3 was two-fold, first to validate and standardize the format of output from Task 2, and then second to begin to collect the creative content in the form of a persona. For example, we used Task 3 to transition from \textit{Sneaky Thieves who stole the gold} to \textit{Sneaky Thieves} to \textit{Sneaky Thief}. Based on worker feedback from initial pilots, we found that balancing creative and mechanical work in the same task kept workers more engaged with the tasks at hand.

The most common mistake that surfaced in the initial pilots was incomplete entries for tasks that didn't actually require correction, for example if the provided form was simply \textit{Traveler}. We chose to embrace this format and assume that unfilled entries were already in their base form. The second most common mistake was describing personas from a third person perspective. This occurrence required manual filtering, as in some cases it was actually somewhat character appropriate to have a persona in that format, such as for an uneducated goblin. We filtered out a majority of these by searching for direct overlap between the provided character name and the persona. Ultimately it's easy to extract the examples that have the clearest grounding format by filtering for examples that contain \textit{"I"}, so as these examples provide more variety in the dataset we chose to keep them.

A remaining issue brought forth by our singular-form constraint is that it was somewhat ambiguous how one would get the singular form of a collective term such as \textit{family}. In most cases we found that workers would choose to provide the format of \textit{collective member} or simply \textit{person}, which sometimes led to vague personas and thus less strong grounding in followup tasks. The content is still workable in these cases though, just not as ideal as we might have wanted. A possible route for improvement here would be a task that asks workers to create a few possible members for a collective for any character we currently have annotated as a member. It is important to note that these cases account for just 44 out of the 1755 collected characters.

One issue of note that surfaced in Task 4 was that workers occasionally described clothing that would potentially lead to risky actions and conversation material, so we chose to eliminate undergarments from the dataset to prevent the creation of inappropriate combinations with the \textit{remove} action. This was included as something to not write about in the task text.

\paragraph{Objects}

\begin{figure*}
    \centering
    \includegraphics[width=\textwidth]{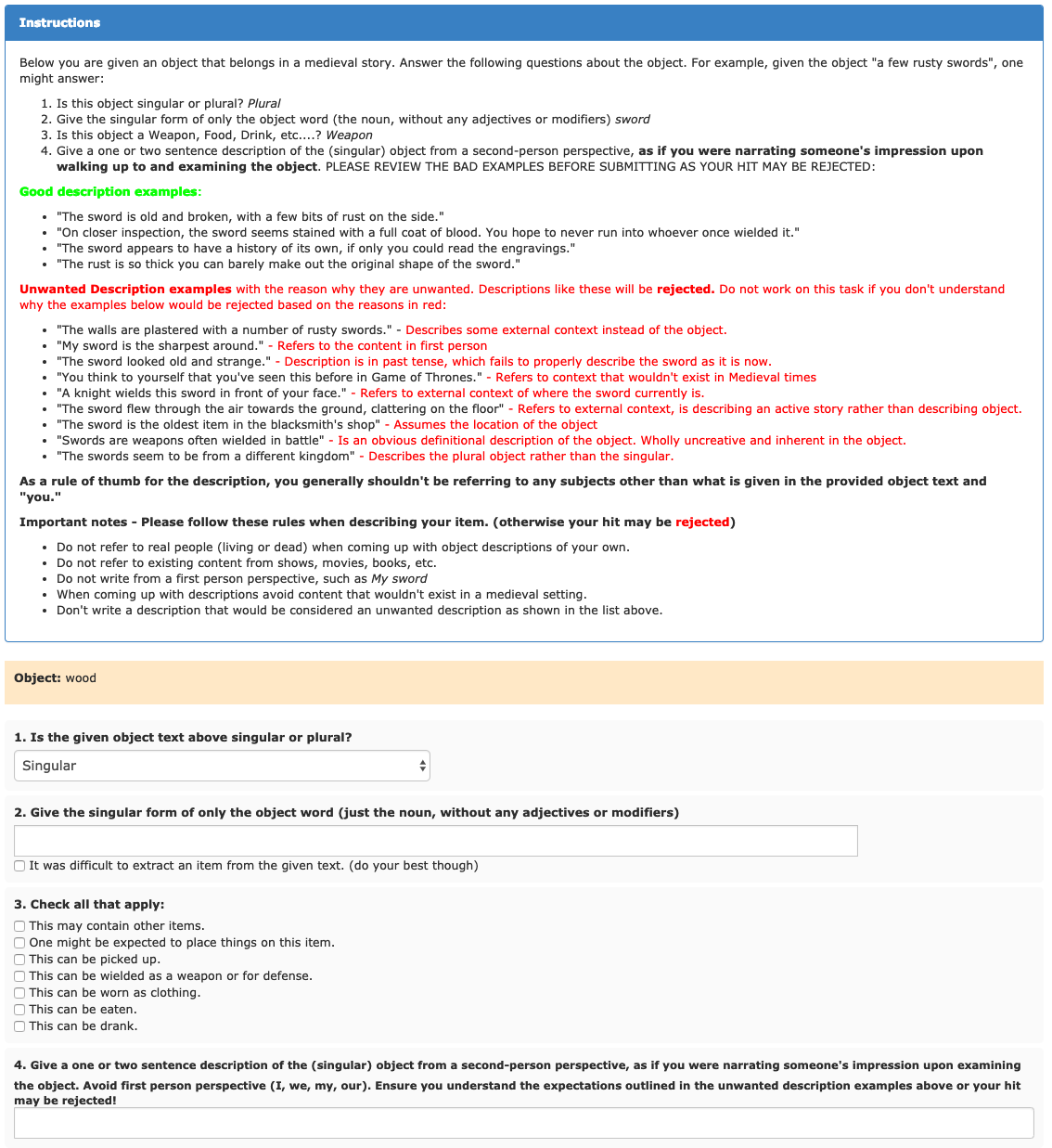}
    \caption{Form for Crowdsourcing Task 5}
    \label{fig:turk-task-5}
\end{figure*}

The object task is most similar to Task 3, but refocused on annotating objects that were specified in Tasks 2 and 4. It took a step to correct the provided span and give a textual description of the object. It also asked for a number of affordances, namely if the object can be picked up, is a container, is a surface, can be eaten, can be drank, can be worn, or can be wielded. We also collected a flag for if a particular example was not appropriate for the dataset or was hard to make sense of. This content was also collected as a form-based task, and we refer to it as \textit{Task 5} and display it in Figure \ref{fig:turk-task-5}. We use complete sentence validation on the text description as a simple quality filter as in previous tasks.

The methodology for Task 5 is very similar to Task 3, trying to both standardize data from previous tasks and act as a filter for bad content that could have been overlooked before. It similarly had both a mechanical data entry and creative component, which tried to keep engagement up.

Overall the largest problem that was surfaced in the pilots was that workers tended to come up with descriptions for objects that were incompatible with our long term goal of having modular components that can be mixed and matched between rooms and scenarios. This came up in many forms, such as workers describing objects as if they were being used in a scene happening in the present, as in \textit{the sword glimmered in the hands of the knight, wielded high in the sky in a call to battle}. While creative, these ultimately were not what we were looking for, so we explicitly called out descriptions like this and many others as being undesired content in our task description. We then manually checked a few examples from each worker to ensure that the data coming in for the final task mostly adhered to this rule.

It is important to note that the object affordances collected are somewhat noisy due to different possible interpretations of the primary object or the tags. Something like a \textit{branch} could be valid as a surface in one circumstance, or a gettable weapon in another. We attempted to reconcile some individual affordances where the pairings of affordances didn't make much sense (for example, very few objects should be both a weapon and edible). This helped with certain objects that were over-tagged, however we haven't used any methods for reconciling scenarios where an object was under-tagged.

\paragraph{Dialogues}

\begin{figure*}
    \centering
    \includegraphics[width=\textwidth]{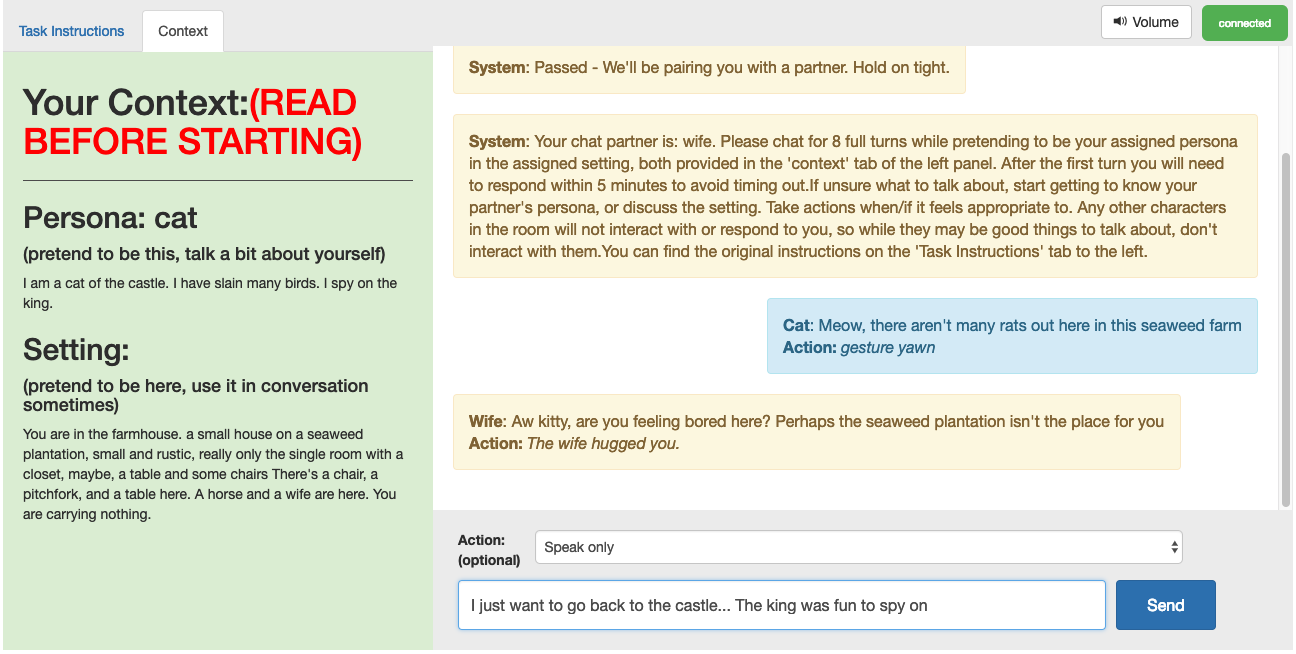}
    \caption{Chat Interface for Crowdsourcing Task 6}
    \label{fig:turk-task-6}
\end{figure*}

Dialogue collection was the hardest task to get correct, and required the largest number of pilot tasks and worker quality control techniques to get to a place that we were satisfied with. The final approach included creating a simple but deliberate onboarding test that needed to be passed in order to move forward with the task at all, collecting mutual feedback from workers about each other, setting timeouts for how quickly workers needed to respond to each turn, and manually validating a few examples from each worker. Each of these steps aimed to solve a different problem, as described in the rest of this section. We will refer to this task as \textit{Task 6}, and it was collected using the ParlAI-MTurk interface as shown in Figure \ref{fig:turk-task-6}.

Firstly, we needed to pair two workers together in order to properly collect dialogues with people playing two different roles without necessarily having insider information into the decisions of each others' turns. While pairing workers solves this problem, it makes the worker experience incredibly dependent on the quality of the worker that they are paired with you. Furthermore, if a worker is paired with a worker that is extremely low quality, the whole dialogue may need to be discarded or is otherwise only useful as an example for how a model might want to react to bad input. If the other worker is good, this makes having any bad workers in the pool not just a poor experience for workers but expensive for the collection process in general. This is the problem that the initial onboarding test aimed to solve. The requirements for passing included entering a specific correct answer as well as at least 4 characters of into the text field. The required action was created such that a worker would have to read and understand the provided persona and setting, how the two interact, the characters and actions available, and be able to synthesize all of the information with an understanding of how to use the interface to send the correct answer. The test required getting the single action correct in 3 attempts. Failing the test on any attempt would permanently soft block a worker from working on Task 6 in the future.

The above test did a lot of work for flagging workers that were well below the bar for completing Task 6 at the level we wanted for the dataset, however as it was a one turn test and it had no way to fully evaluate the quality by which workers would actually incorporate their persona and the setting into their dialogue turns. Furthermore, it didn't filter out workers that would take too much time on their turns and thus cause their partners to disengage and provide lower quality responses, potentially due to working on other tasks in the background and doing too much context switching. We solved these problems separately.

In order to handle low quality workers, we allowed workers the opportunity to rate each other at the end of each dialogue, and to provide tags about the experience. We found that positive feedback was generally noisy and hard to get signal from, but negative feedback almost always correlated to a worker who was providing bad content. As a bonus, workers gave us positive feedback about this capability, as it allowed them to filter out workers that made the task less engaging and interesting for them. We reviewed this feedback periodically while tasks were running and soft-blocked workers low quality workers whenever they were flagged.

In order to handle the influence of response time on task quality, we set a maximum response time of 5 minutes for any given turn, and overall started soft blocking workers that were consistently above 2 minutes for each message, even if their particular content was pretty good. This improved collection times and did not seem to negatively affect quality.

After this point, manually checking the collected conversations still surfaced a few bad examples when viewing one chat per worker rather than arbitrarily sampling the dataset. In order to remedy this, the last quality check was a direct evaluation of at least 2 dialogues from each worker. This caught a few overlooked instances from workers that didn't necessarily work on enough tasks to get flagged by one of our consistently reviewing workers. Generally this surfaced some quality issues surrounding profanity, inappropriate content for the given setting, and entire misunderstanding of the task at hand such as never using the persona or location as grounding context in the conversation. As not all workers were particularly diligent raters (as confirmed by the low signal of positive ratings - workers don't necessarily want to flag each other as bad), a few workers were able to slip through the cracks up until this point due to not completing enough tasks to encounter a rater that flagged them.

One small acknowledgement throughout the dialogues is that there are still misspellings, improper grammar, mistaken keystrokes, and such. While the rate of occurrence is orders of magnitude lower than we observed in the initial pilots, it is hard to separate cases where it is a genuine mistake versus cases where it is appropriate for the character, such as a \textit{pirate} using seaworthy lexicon and adding extra R's to suggest a pirate-like drawl, or a \textit{snake} that slips in extra S's to better play the role. 

\section{Descriptions of Actions and Emotes}
\label{appendix:detailed-actions-emotes}

The LIGHT action set builds upon the graph framework introduced in Mastering the Dungeon \cite{yang2017mastering}. The basic idea presented is that everything in the text adventure game can be represented as nodes, and then state is described by edges between those nodes. In this way, an agent and an object can be in a room, and that agent can be carrying a different object or a container might have an object inside as well by the same kind of relation. After defining this relationship, we can further define a set of actions that can be taken based on a combination of the state of the graph and the attributes of nodes in that graph. The available actions for the dialogues collected in this dataset, along with the constraints for applying those actions, are available in Table \ref{table:light_actions}. We used the crowdsourced object affordances to set the correct attributes for nodes in the graph (if the object can be picked up, is a container, is a surface, can be eaten, can be drank,can be worn, or can be wielded). 

For the emotes, we paired down a list of emotes sourced from existing MUDs to reduce redundancy and task complexity at the acknowledged cost of expressiveness. This led us to select just one out of \textit{scream}, \textit{shout}, and \textit{yell} instead of keeping them all, as having all of the emotes would lead to a more complicated crowdsourcing task than we wanted to risk. We ended up with a set of 22 emotes, listed in Figure \ref{appendix:emote-list}.

\begin{figure}[t]
\begin{center}
\scalebox{0.88}{
\begin{tabular}{l}
\toprule
applaud, blush, cry, dance, frown, gasp, grin, groan, \\
growl, laugh, nod, nudge, ponder, pout, scream,\\ 
shrug, sigh, smile, stare, wave, wink, yawn \\
\bottomrule
\end{tabular}
}
\caption{
Emote options within the LIGHT platform
\label{appendix:emote-list}
}
\end{center}
\end{figure}

\begin{table*}[t]
\begin{center}
\small
\begin{tabular}{l|ll}
 \toprule
\textbf{Action} & Constraints & Outcome \\
\midrule                
get \textit{object} & actor and \textit{object} in same room & actor is carrying \textit{object}  \\
 & \textit{object} is gettable & \\
\midrule                
drop \textit{object} & actor is carrying \textit{object} & \textit{object} is in room \\
 & \textit{object} is gettable & \\
\midrule
get \textit{object1} from \textit{object2} & Actor and \textit{object2} in same room & actor is carrying \textit{object1}  \\
 & \textit{object1} is gettable & \\
 & \textit{object2} is surface or container & \\
 & \textit{object2} is carrying \textit{object1} & \\
\midrule
put \textit{object1} in/on \textit{object2} & Actor and \textit{object2} in same room & \textit{object2} is carrying \textit{object1}  \\
 & \textit{object2} is container or surface& \\
 & actor is carrying \textit{object1} & \\
\midrule
give \textit{object} to \textit{agent} & Actor and \textit{agent} in same room & \textit{agent} is carrying \textit{object} \\
 & \textit{object} is a member of actor & \\
\midrule
steal \textit{object} from \textit{agent} & actor and \textit{agent} in same room & actor is carrying \textit{object} \\
 & \textit{object} is a member of \textit{agent} & \\
\midrule
hit \textit{agent} & Actor and \textit{agent} in same room & inform \textit{agent} of attack \\
\midrule
hug \textit{agent} & Actor and \textit{agent} in same room & inform \textit{agent} of hug \\
\midrule
drink \textit{object} & actor is carrying \textit{object} & inform actor of drinking successfully \\
& \textit{object} is a drink & \\
\midrule
eat \textit{object} & actor is carrying \textit{object} & inform actor of eating successfully \\
& \textit{object} is a food & \\
\midrule
wear \textit{object} & actor is carrying \textit{object} & actor is wearing \textit{object} \\
& \textit{object} is wearable & \\
\midrule
wield \textit{object} & actor is carrying \textit{object} & actor is wielding \textit{object} \\
& \textit{object} is a weapon & \\
\midrule
remove \textit{object} & actor is wearing/wielding \textit{object} & actor is carrying \textit{object} \\
& \textit{object} is wearable or a weapon & \\
\bottomrule
\end{tabular}
\caption{LIGHT actions and constraints
\label{table:light_actions}
}
\end{center}
\end{table*}

\section{Descriptions of Human Evaluations}
\label{appendix:human-eval-stats}

As crowdworkers can sometimes be inconsistent, we set up two filters to onboard workers into being fair representatives for human perfomance on the task. The first gave workers a few chances to select the correct input for a turn each of dialogue, emote, and action on a scenario we created to strongly hint at the correct answer. We then chose to use performance on the training set as a secondary filter to have workers that were capable of the task. Each of the tasks has a different level of difficulty, so we selected reasonable benchmark values based on our own performance on the tasks. For dialogue, this required getting all 7 of the turns from the training set correctly. For actions, this required getting 6 out of 8 turns from the training set correctly. Lastly for emoting, we required getting only 2 out of 8 turns from the training set correctly. On the seen set, our accuracy on the dialogue, action, and emote tasks were calculated from 217, 165, and 211 turns respectively. On the unseen set, we calculated the accuracy from 196, 114, and 209 turns respectively. 

\section{Embedding Visualizations}
\label{app:tsne-etc}

To explore the diversity of LIGHT, we use t-SNE \cite{Maaten2008VisualizingDU} to visualize the embeddings of the different atomic dataset elements -- locations, objects, characters, and actions. We use two different embeddings methods to tease out two key aspects of our dataset: 1) the \textit{interconnectedness} of grounding information (relationships between different types of elements, such as the actions available around given objects, or in a given location), and 2) \textit{coverage} (the variety of different objects, locations, and characters in our world). 

\begin{figure*}
    \centering
    \includegraphics[width=\textwidth]{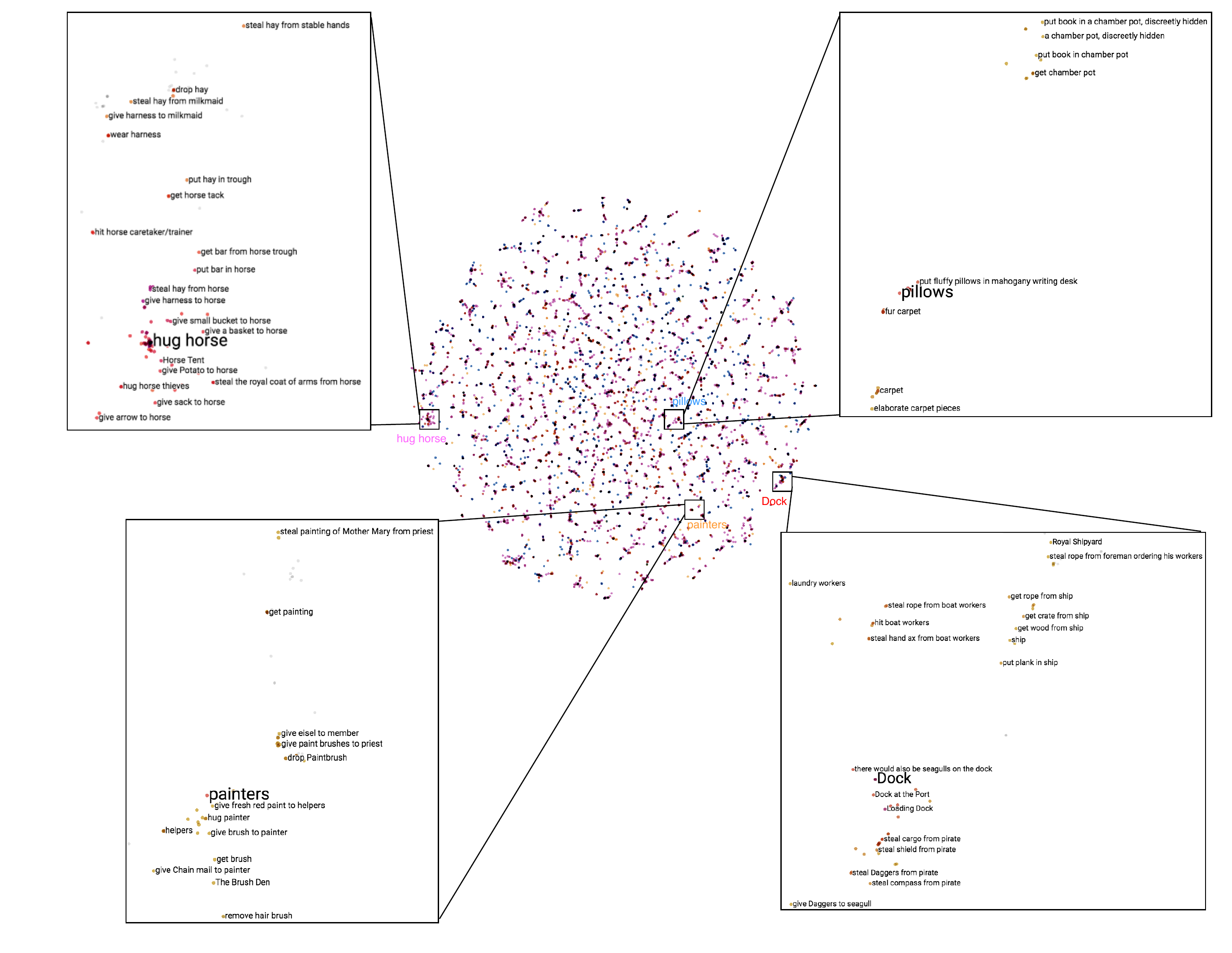}
    \caption{t-SNE Visualization of Starspace embeddings learned directly from the LIGHT Dataset. Color denotes each element type, either location, character, action, or object. We select four neighborhoods to explore, for each of the base element types: ``Dock'' (location), ``painters'' (character), ``hug horse'' (action), and ``pillows'' (object).}
    \label{fig:starspace-embeddings}
\end{figure*}

To explore the \textit{interconnectedness} of our dataset, we visualize the embeddings learned when training the baseline Starspace ranking model on the task of dialogue, action, and emote prediction, in this case with no pretrained vectors so learning comes from our dataset alone. The t-SNE visualizations of these Starspace embedding can be found in Figure \ref{fig:starspace-embeddings}. Because the Starspace model operates by mapping all inputs and outputs to a shared embedding space, we find the learned embeddings capture many of the nuances and relationships between different elements of our dataset. For example, looking at the nearest neighbors for the location ``Dock'' (the bottom-right of Figure \ref{fig:starspace-embeddings}), we see actions like ``get crate from ship,'' ``put plank in ship,'' objects like ``ship'' and ``rope,'' and characters like ``boat workers.'' We see similar relationships captured when looking at nearest neighbors for the ``painters'' characters, the ``hug horse'' action, and the ``pillows'' objects.

\begin{figure*}
    \centering
    \includegraphics[width=\textwidth]{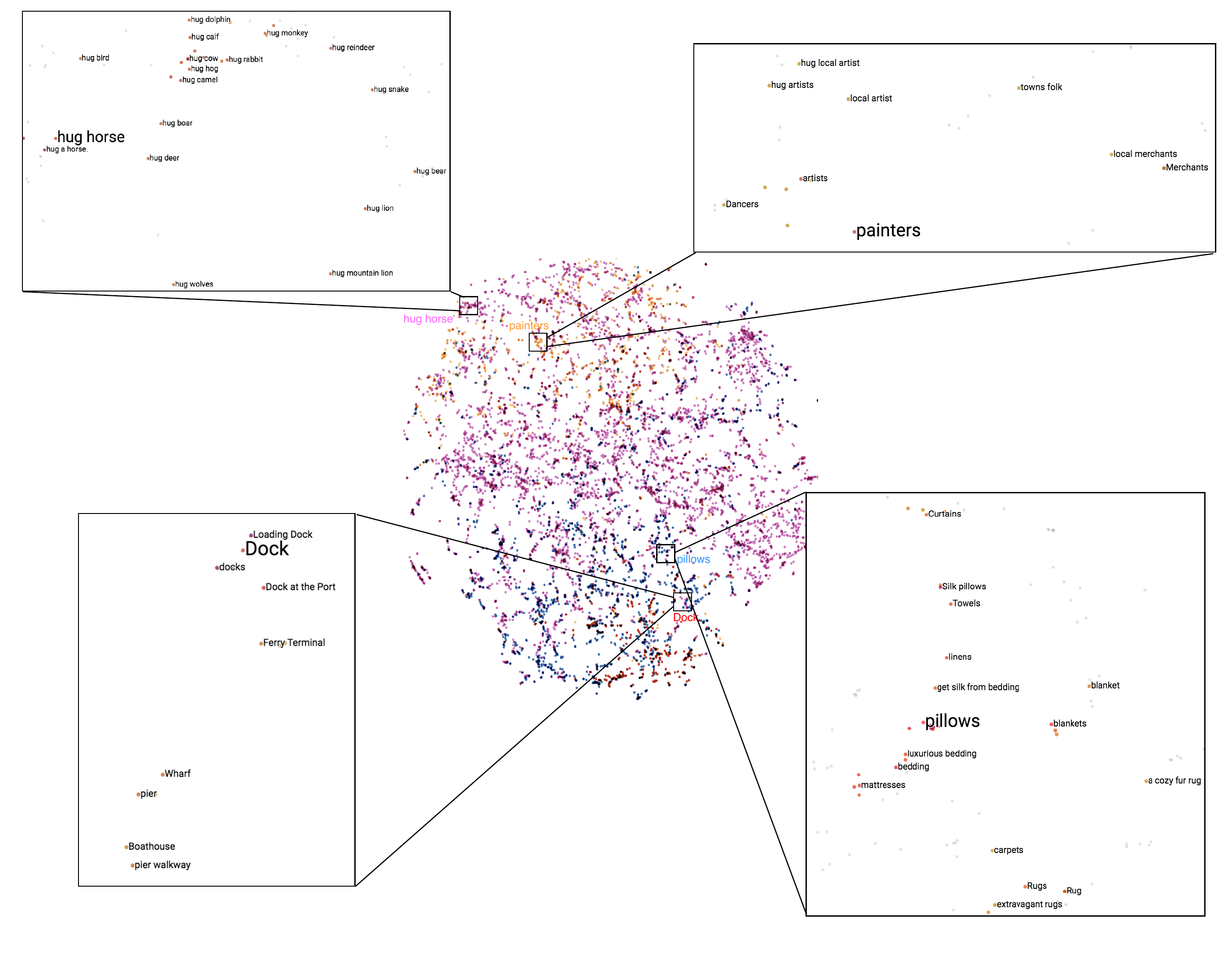}
    \caption{t-SNE Visualization of pretrained GLoVe embeddings for different LIGHT elements. Color denotes each element type, either location, character, action, or object. We select four neighborhoods to explore, for each of the base types: ``Dock'' (location), ``painters'' (character), ``hug horse'' (action), and ``pillows'' (object).}
    \label{fig:glove-embeddings}
\end{figure*}

To explore the \textit{coverage} of our dataset, we use pretrained GLoVe word embeddings \cite{Pennington2014GloveGV}, trained on the Common Crawl corpus. As each dataset element can consist of multiple words (e.g. ``give the horse a potato,'' or ``The Queen's Chamber''), we take the mean of the GLoVE vectors for each word as the fixed vector embedding for the element. The t-SNE visualizations of these GLoVe-embedded elements can be found in Figure \ref{fig:glove-embeddings}. Unlike the Starspace embeddings, which capture the structure present in the relationships between different types of dataset elements, we find that the GLoVe embeddings capture the breadth and semantic similarities of dataset elements. For example, looking at the nearest neighbors for the embedding of the ``Dock'' location, we see similar locations present in our dataset, like ``Ferry Terminal,'' ``Wharf,'' ``pier,'' and ``Boathouse.'' Similarly, if we look at the nearest neighbors for the ``pillows'' objects, we see other objects like ``bedding,'' ``mattresses,'' ``rugs,'' ``towels,'' and ``curtains.''

\section{Action and Emote Relationships} \label{appendix-sec:location-categories}

\begin{figure*}
\centering
\begin{subfigure}{0.49\textwidth}
    \centering
    \label{fig:act-act}
    \includegraphics[width=\linewidth]{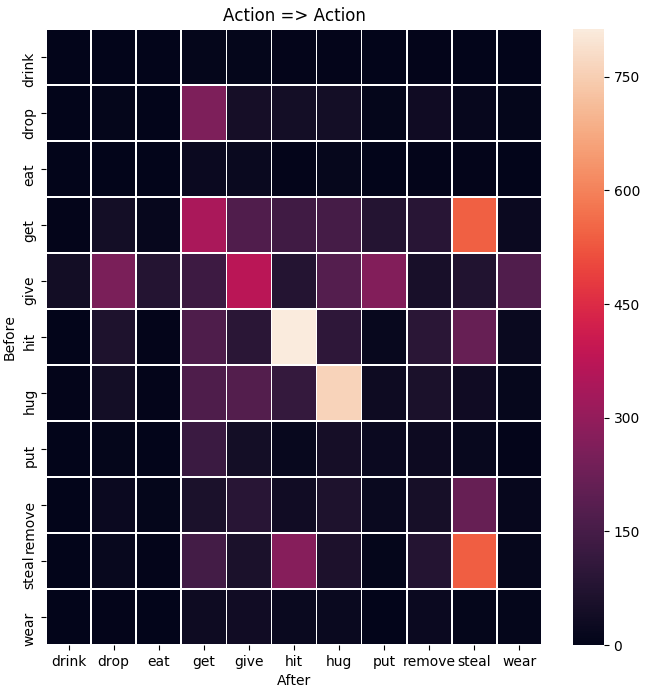}
\end{subfigure}
\begin{subfigure}{0.49\textwidth}
    \centering
    \label{fig:act-emote}
    \includegraphics[width=\linewidth]{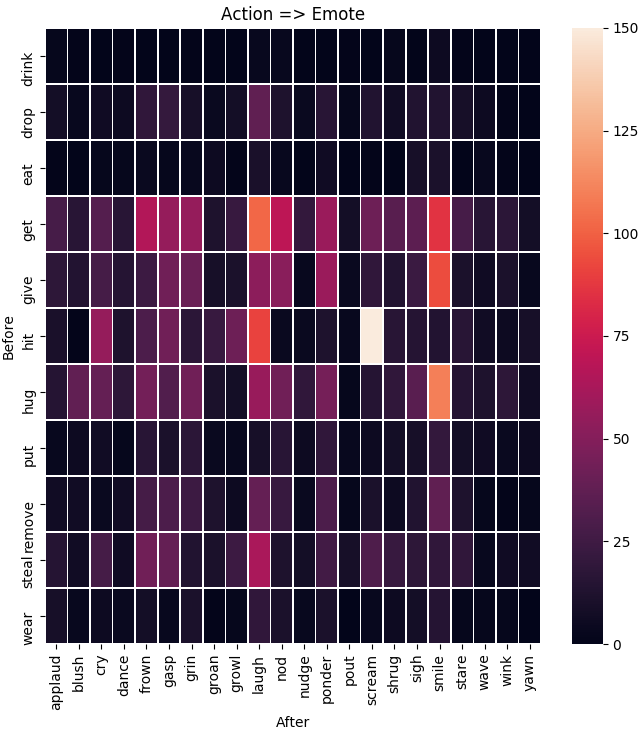}
\end{subfigure} \\
\vspace{2cm}
\begin{subfigure}{0.49\textwidth}
    \centering
    \label{fig:act-act}
    \includegraphics[width=\linewidth]{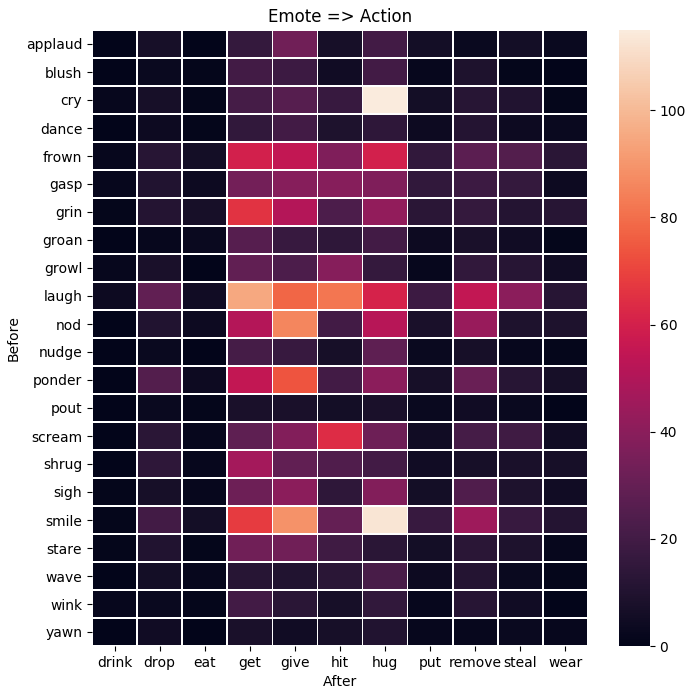}
\end{subfigure}
\begin{subfigure}{0.49\textwidth}
    \centering
    \label{fig:act-emote}
    \includegraphics[width=\linewidth]{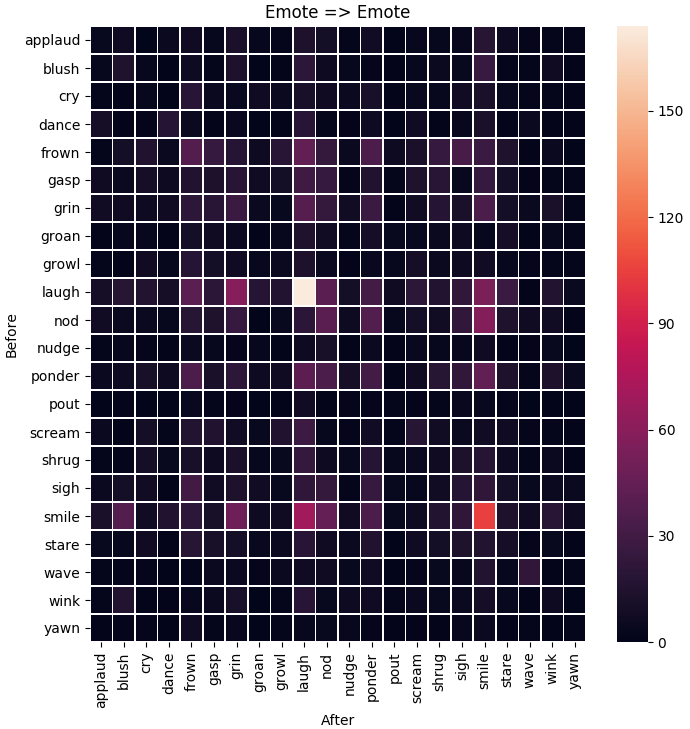}
\end{subfigure}
\caption{Heatmaps displaying causal relationships between Emotes and Actions. LIGHT is emotionally diverse -- there are many different ways for a character to respond to another's emotional state. However, there are a few strong trends present: screaming or hitting someone back after being hit, laughing together, and comforting a crying character with a hug.}
\label{fig:heatmaps}
\end{figure*}

To visualize the interaction trends between actions and emotes in LIGHT, we present heatmaps (in Figure \ref{fig:heatmaps}) counting the number of occurrences of each immediately before or after one's partner performs an action or emote. While responses to an action or emote can evolve over multiple timesteps, we limit this visualization to action relationships within a single timestep. Additionally, to effectively measure trends in physical actions, we cluster all physical actions by the root word (for example, ``steal the sword from the soldier'' becomes ``steal''). 

While for the most part there are a multitude of different observed physical and emotional responses for each partner move, there are certain interesting trends to observe. Looking at the top-left of Figure \ref{fig:heatmaps}, we see that if one's partner makes a ``hit'' action, the most likely response is to ``hit'' back. Looking at the same plot, we see that ``hug'' actions are similarly reciprocated. If we look at the interplay between physical actions and emotes (top-right of Figure \ref{fig:heatmaps}) we see a relationship between one's partner taking a ``hit'' action, and issuing a ``scream'' emote in response. Going the other direction and looking at the relationship between emotes and physical actions, we see that performing a ``cry'' or ``smile'' emote is likely to be met with either a consoling or celebratory ``hug.'' Finally, looking at the relationships between a partner's emote and an emote response, we see that positive emotes like ``laugh'' and ``smile'' are likely to be reciprocated with a similar (if not identical) emote.

\begin{table*}[t]
\begin{center}
\scalebox{0.88}{
\begin{tabular}{ll}
\toprule
{\bf Seen} & Abandoned, Bazaar, Cave, Countryside, Desert, Dungeon, Farm, Forest, Graveyard, Inside Castle, \\
& Inside Church, Inside Cottage, Inside Palace, Inside Temple, Inside Tower, Jungle, Lake, Mountain, \\
& Outside Castle, Outside Church, Outside Cottage, Outside Palace, Outside Temple, Outside Tower, \\
&  Port, Shore, Swamp, Tavern, Town, Trail, Wasteland\\
{\bf Unseen} & City in the Clouds, Frozen Tundra, Magical Realm, Netherworld, Supernatural, Underwater Aquapolis\\
\bottomrule
\end{tabular}
}
\caption{
Location categories for both the seen and unseen sets of locations.
\label{appendix:location-categories}
}
\end{center}
\end{table*}

\begin{figure*}[t]
  \begin{center}
    \begin{footnotesize}
      \begin{tabular}{l|l}
        \toprule
        \textbf{Persona 1: A serving wench} & \textbf{Persona 2: Cleaning person} \\
        \midrule
        I work at the local tavern. & I scrub the palace floors day and night. \\
        I enjoy talking to the soldiers that frequent the tavern. & My bones are brittle from the hard labor, \\
        I steal tips from the change of the patrons. & but my heart is strong. \\
         & I save my day's coin, never spending it. \\
         & I am frugal and creative. \\
         & I long for the day when I can buy my freedom \\
         & from the Queen. \\
         & It has been 40 years, but I am patient. \\
        \midrule
        Carrying: a Wine, a purse, a plate & Carrying: a rag, a Broom, a Bucket \\
Wearing: a polishing cloths, a ring, a scarves, a dress, a cloth & Wearing: a boot \\
        \midrule
        \multicolumn{2}{l}{\textbf{{Setting:}} The kitchen tavern is a small cramped room, with wooden cabinets and surfaces made out of stone} \\
        \multicolumn{2}{l}{tiles. There are many bottles of liquors and beers on the shelves, and there are buckets full of ice and} \\
        \multicolumn{2}{l}{other things. There is one shelf full of food items. There is a basin for water, and a bunch of knives} \\
        \multicolumn{2}{l}{and tools in a drawer.} \\

        \bottomrule

        \multicolumn{2}{l}{\textbf{A serving wench:} Hello - how are things going so far?} \\
        \multicolumn{2}{l}{\textit{~~Action:} get food item} \\
        \multicolumn{2}{l}{\textbf{Cleaning person:} Ah, sorry, miss.  I know I look a fair bit like me sister who works hereabouts, but I'm actually Gretchen,} \\
        \multicolumn{2}{l}{the sister that works up in the palace.  Perhaps she might 'ave mentioned me?} \\
        \multicolumn{2}{l}{\textit{~~Action:} gesture smile} \\
        \multicolumn{2}{l}{\textbf{A serving wench:} You're Gretchen? I've heard so much about you! Hello and welcome. } \\
        \multicolumn{2}{l}{\textit{~~Action:} gesture smile} \\
        \multicolumn{2}{l}{\textbf{Cleaning person:} The Queen's in a bit of a state, what with the King being a bit more open with his mistress,} \\
        \multicolumn{2}{l}{so I thought I'd just hide hereabouts for a spell.  I'd be happy ta lend a hand, though!} \\
        \multicolumn{2}{l}{\textbf{A serving wench:} Oh no! Is he really carrying on like that again? We'd appreciate any help you can give. Thanks!} \\
        \multicolumn{2}{l}{\textit{~~Action:} get bucket} \\
        \multicolumn{2}{l}{\textbf{Cleaning person:} Seems like a right busy crowd out there!  'as it been a bit on the noisier side?  } \\
        \multicolumn{2}{l}{\textit{~~Action:} put rag in basin for water} \\
        \multicolumn{2}{l}{\textbf{A serving wench:} Heavens, yes! It'll only get rowdier as the day goes on into night. Lots of bourbon and shots you know. } \\
        \multicolumn{2}{l}{\textit{~~Action:} gesture laugh} \\
        \multicolumn{2}{l}{\textbf{Cleaning person:} Ach, I don't think I'll ever be gettin the stains outta this rag, but it'll do! } \\ 
        \multicolumn{2}{l}{Do ya make much in the way of coins here?  Can always use an extra bit o'coin, eh?} \\
        \multicolumn{2}{l}{\textit{~~Action:} get rag from basin for water} \\
        \multicolumn{2}{l}{\textbf{A serving wench:} You can, especially if you take some from the change of the patrons.} \\
        \multicolumn{2}{l}{They're so drunk they never catch it! } \\
        \multicolumn{2}{l}{\textit{~~Action:} put Wine in cabinet} \\
        \multicolumn{2}{l}{\textbf{Cleaning person:} O-oh?  Is that.. well I suppose if they've enough coin ta spend on this, then a coin 'ere} \\
        \multicolumn{2}{l}{or there won't go amiss.} \\
        \multicolumn{2}{l}{\textit{~~Action:} gesture ponder} \\
        \multicolumn{2}{l}{\textbf{A serving wench:} Exactly. That's what I say. So, are you just here for a day?} \\
        \multicolumn{2}{l}{\textbf{Cleaning person:} Well, I suppose it all depends on how the King takes to 'is wife destroying his prized tapestry from the}\\
        \multicolumn{2}{l}{last war. Sometimes he has a short temper, that one.} \\
        \multicolumn{2}{l}{\textit{~~Action:} gesture nod} \\
        \multicolumn{2}{l}{\textbf{A serving wench:} Sounds like his wife may, too. } \\
        \multicolumn{2}{l}{\textit{~~Action:} gesture laugh} \\
        \multicolumn{2}{l}{\textbf{Cleaning person:} Aye, ye don't know the half of it.  I feel like most of me days are spent sweeping up broken crockery}\\
        \multicolumn{2}{l}{ and china than the usual dusting that I'm supposed ta be doin'!} \\
        \multicolumn{2}{l}{\textit{~~Action:} gesture sigh} \\
      \bottomrule
      \end{tabular}
      \caption{Example dialogue from the {\sc LIGHT} dataset.
  \label{appendix:dialogues-first}}
    \end{footnotesize}
  \end{center}
\end{figure*}

\begin{figure*}[t]
  \begin{center}
    \begin{footnotesize}
      \begin{tabular}{l|l}
        \toprule
        \textbf{Persona 1: Boar} & \textbf{Persona 2: Faery} \\
        \midrule
        I am an ugly animal. & I am a faery, one of the fae. \\
        I am eaten sometimes for food. & We are magical people who live in the forest. \\
        I do not like those who try to hunt me. & We try to avoid humans, because they will catch and enslave \\
         &  us for their own use, if they can. \\
         & Our magickal skills enable us to live comfortable lives, and to \\
         & keep away from those who would do us harm. \\
        \midrule
        Carrying: \textit{nothing} & Carrying: \textit{nothing}  \\
Wearing: \textit{nothing} & Wearing: a glittery pointed cap, a Earring \\
        \midrule
        \multicolumn{2}{l}{\textbf{{Setting:}} The entryway to the brush den is made of arched bushes and various plants, as is the ceiling and} \\
        \multicolumn{2}{l}{walls. The inside is furnished with seats made of plant roots that have grown together, and a table made} \\
        \multicolumn{2}{l}{of wood adorned with flowers and wooden cups and plates. A small vine dangles with a glowing floor from} \\
        \multicolumn{2}{l}{the middle of the ceiling that lights the room dimly. Three small beds lay in one corner of the room made} \\
        \multicolumn{2}{l}{of interlaced grass and leaves.} \\

        \bottomrule

        \multicolumn{2}{l}{\textbf{Boar:} Hello faery, do you think you could help me?} \\
        \multicolumn{2}{l}{\textbf{Faery:} A talking boar! You must be enchanted by the witch. How can I help you, my dear friend?} \\
        \multicolumn{2}{l}{\textbf{Boar:} That is correct, I am enchanted... or cursed rather... by an evil witch. That is why I need help.} \\
        \multicolumn{2}{l}{\textbf{Faery:} I suspected as much. Please, tell me more so that I may help you.} \\
        \multicolumn{2}{l}{\textit{~~Action:} gesture frown} \\
        \multicolumn{2}{l}{\textbf{Boar:} Thank you so much! I hate it here in this Den! } \\
        \multicolumn{2}{l}{\textit{~~Action:} hug faery} \\
        \multicolumn{2}{l}{\textbf{Faery:} Are you being held prisoner in this den? These vines have grown about everywhere - it's impossible to get around.} \\
        \multicolumn{2}{l}{\textbf{Boar:} No, this is the only place I can live without being hunted by the humans. } \\
        \multicolumn{2}{l}{\textbf{Faery:} Oh, I see. I hate their type. They are always trying to catch and enslave me. You say you are cursed by the witch.}\\\multicolumn{2}{l}{I can help you but I need to know what curse she has put upon you.} \\
        \multicolumn{2}{l}{\textbf{Boar:} Well now that you say you hate them, I don't know if I should tell you...} \\
        \multicolumn{2}{l}{\textit{~~Action:} gesture frown} \\
        \multicolumn{2}{l}{\textbf{Faery:} Suit yourself, my friend. You would hate them, too, if they were constantly enslaving your people.} \\
        \multicolumn{2}{l}{\textit{~~Action:} hug boar} \\
        \multicolumn{2}{l}{\textbf{Boar:} Well I am a human...} \\
        \multicolumn{2}{l}{\textbf{Faery:} You! A human?! Perhaps we can work together. I can help return you to your form, and you can prevent your people}\\
        \multicolumn{2}{l}{ from capturing faeries like me!} \\
        \multicolumn{2}{l}{\textit{~~Action:} gesture gasp} \\
        \multicolumn{2}{l}{\textbf{Boar:} I think we can do that, I used to be quite the ruler when I was human.} \\
        \multicolumn{2}{l}{\textbf{Faery:} Excellent. Let me speak the magic words and enchant this flower. Then you can eat it and in three days you}\\
        \multicolumn{2}{l}{will be human again!} \\
        \multicolumn{2}{l}{\textit{~~Action:} get Flower} \\
      \bottomrule
      \end{tabular}
      \caption{Example dialogue from the {\sc LIGHT} dataset.
  \label{appendix:dialogues-second}}
    \end{footnotesize}
  \end{center}
\end{figure*}

\begin{figure*}[t]
  \begin{center}
    \begin{footnotesize}
      \begin{tabular}{l|l}
        \toprule
        \textbf{Persona 1: President} & \textbf{Persona 2: Mayor} \\
        \midrule
        I won the election. & I am the mayor of the village. \\
        People listen to what I say. & I help the king keep order of the subjects. \\
        I am very powerful. & I have a high position in the kingdom. \\
        \midrule
        Carrying: a book & Carrying: a document, a key \\
Wearing: a crown & Wearing: a jewelry, a ceremonial hat \\
        \midrule
        \multicolumn{2}{l}{\textbf{{Setting:}} Large and extravagant, the room is adorned with crystals, polished gold and sapphires. There's long} \\
        \multicolumn{2}{l}{tables with beautiful silk table clothes covering them. Plush chairs line the tables. In front of each} \\
        \multicolumn{2}{l}{table is plates made from fine China, next to the plates is sterling silver silverware laid upon a pure} \\
        \multicolumn{2}{l}{white napkin. There's a stage where there's 2 thrones overlooking the table. The whole ceiling is large} \\
        \multicolumn{2}{l}{and adorned with chandeliers and garnished with gold accents.} \\

        \bottomrule

        \multicolumn{2}{l}{\textbf{President:} So much luxuty in this room, many rulers have been here before us. } \\
        \multicolumn{2}{l}{\textbf{Mayor:} This is a very luxurious room, President. Here. The King told me to pass this on to you.} \\
        \multicolumn{2}{l}{\textit{~~Action:} give document to president} \\
        \multicolumn{2}{l}{\textbf{President:} This is a letter for the king assuring my rule in this part of the kingdom, thank you mayor,}\\
        \multicolumn{2}{l}{ I will place document on a sacred place} \\
        \multicolumn{2}{l}{\textit{~~Action:} put document in polished gold} \\
        \multicolumn{2}{l}{\textbf{Mayor:} He also gave me this to give to you. He told me that you need to keep this in a VERY safe place until he comes for it.} \\
        \multicolumn{2}{l}{\textit{~~Action:} give key to president} \\
        \multicolumn{2}{l}{\textbf{President:} Thats the key to the kingdom I  wonder why the king trusted me with this item, his throne must be in real danger} \\
        \multicolumn{2}{l}{\textbf{Mayor:} Yes, sir. He has also trusted me with other objects such as this to keep. We are his trusted allies.} \\
        \multicolumn{2}{l}{\textbf{President:} Thank you, he must trust you a lot as well, here take this as a sign of my affection, its a royal jewel taken out of the}\\
        \multicolumn{2}{l}{ volcano of Mordor} \\
        \multicolumn{2}{l}{\textit{~~Action:} get sapphire} \\
        \multicolumn{2}{l}{\textbf{Mayor:} This is absolutely beautiful. I have heardd that Mordor is beyond the wall. Is that true?} \\
        \multicolumn{2}{l}{\textit{~~Action:} remove ceremonial hat} \\
        \multicolumn{2}{l}{\textbf{President:} That is true, only the bravest warriors go to that place, its full with dangers and fierce animals} \\
        \multicolumn{2}{l}{\textbf{Mayor:} Oh dear. What if our King went there? What if there is something there he needs for battles to come!} \\
        \multicolumn{2}{l}{\textit{~~Action:} wear ceremonial hat} \\
        \multicolumn{2}{l}{\textbf{President:} Our king is a fierce warrior but I am worried, who knows what can happen if he goes}\\
        \multicolumn{2}{l}{ to a battle in a place like that one} \\
        \multicolumn{2}{l}{\textbf{Mayor:} I heard there are things that walk the forest and come with the cold. We must safe our King!} \\
        \multicolumn{2}{l}{\textbf{President:} Lets hurry then, lets gather an army and go aid our king, heres a book with the names of the bravest soldiers}\\
        \multicolumn{2}{l}{ in the kingdom} \\
        \multicolumn{2}{l}{\textit{~~Action:} give book to mayor} \\
        \multicolumn{2}{l}{\textbf{Mayor:} Oh this book is very amazing. Who is this..Sir Rodryck? } \\
      \bottomrule
      \end{tabular}
      \caption{Example dialogue from the {\sc LIGHT} dataset.
  \label{appendix:dialogues-third}}
    \end{footnotesize}
  \end{center}
\end{figure*}

\begin{figure*}[t]
  \begin{center}
    \begin{footnotesize}
      \begin{tabular}{l|l}
        \toprule
        \textbf{Persona 1: Person} & \textbf{Persona 2: Worms} \\
        \midrule
        I am the maid to the queen. & I am a worm who slides through the dirt. \\
        I get her dressed in the morning and take care of her needs. & I hear many secrets that people tell in the forest. \\
        I live in the servant's quarters on the lower level of the castle. & I can tell some to you, if you would like. \\
        \midrule
        Carrying: \textit{nothing} & Carrying: \textit{nothing} \\
Wearing: an apron & Wearing: \textit{nothing} \\
        \midrule
        \multicolumn{2}{l}{\textbf{{Setting:}} The Winter Gardens' name is a bit of a misdirection - there are flowers here that bloom at many} \\
        \multicolumn{2}{l}{different times of the year. It's almost the size of an entire town square, and it has almost every known} \\
        \multicolumn{2}{l}{flora of the Continent in it.} \\

        \bottomrule

        \multicolumn{2}{l}{\textbf{Person:} Ah, worms are here. They shall be useful for the queen's lizards to feast on.}\\
        \multicolumn{2}{l}{ But first let me remove my apron so as not to dirty it while I collect you all.} \\
        \multicolumn{2}{l}{\textit{~~Action:} remove apron} \\
        \multicolumn{2}{l}{\textbf{Worms:} Noooo! I am a valuable life to preserve! I know things!} \\
        \multicolumn{2}{l}{\textit{~~Action:} gesture scream} \\
        \multicolumn{2}{l}{\textbf{Person:} Worms that can talk!? What sinister magic is this!?} \\
        \multicolumn{2}{l}{\textit{~~Action:} gesture scream} \\
        \multicolumn{2}{l}{\textbf{Worms:} I have been able to talk to humans ever since I was born. } \\
        \multicolumn{2}{l}{\textbf{Person:} How did you acquire such a skill? Do the flowers that bloom in these gardens have special powers that a }\\\multicolumn{2}{l}{simple maid like I cannot understand?} \\
        \multicolumn{2}{l}{\textbf{Worms:} Not the flowers, but out in ther forest i have heard of magical herbs.} \\
        \multicolumn{2}{l}{\textbf{Person:} If not the flowers, then how did you get such powers of speech? I am still curious.}\\\multicolumn{2}{l}{ Surely you did not come all the way from the forest. These gardens are much too large for a simple worm to cross, }\\\multicolumn{2}{l}{even in a thousand lifetimes.} \\
        \multicolumn{2}{l}{\textit{~~Action:} gesture ponder} \\
        \multicolumn{2}{l}{\textbf{Worms:} I have been given this ability from a witch. This is what my father told me.} \\
        \multicolumn{2}{l}{\textbf{Person:} A witch you say? Well then I must surely take you to my queen. }\\\multicolumn{2}{l}{She must know that there is dark magic present in her kingdom.} \\
        \multicolumn{2}{l}{\textbf{Worms:} Oh please no! She will most likely kill me.} \\
        \multicolumn{2}{l}{\textit{~~Action:} gesture gasp} \\
        \multicolumn{2}{l}{\textbf{Person:} Tell me, why should I not take you? Give me a good reason and I may spare you yet.} \\
        \multicolumn{2}{l}{\textbf{Worms:} I know many secrets. I know where stolen goods are.} \\
        \multicolumn{2}{l}{\textbf{Person:} Stolen goods!? Tell me, where they are! I may be able to use them to buy my way out of servitude.} \\
        \multicolumn{2}{l}{\textit{~~Action:} gesture gasp} \\
        \multicolumn{2}{l}{\textbf{Worms:} I heard of these bandits who like to hideout at the tavern by marthas house. }\\\multicolumn{2}{l}{They recently stole gold from the rich oil man.} \\
      \bottomrule
      \end{tabular}
      \caption{Example dialogue from the {\sc LIGHT} dataset.
  \label{appendix:dialogues-fourth}}
    \end{footnotesize}
  \end{center}
\end{figure*}

\begin{figure*}[t]
  \begin{center}
    \begin{footnotesize}
      \begin{tabular}{l|l}
        \toprule
        \textbf{Persona 1: Servant} & \textbf{Persona 2: Court jester} \\
        \midrule
        I come from the lower class. & I am a living joke! my sould is what flies out of your mouth \\
        I do what I am told without question. & when something is funny. \\
        I can not read. & The king hates me!. \\
        I have not seen my family in a long time. &  \\
        \midrule
        Carrying: a rag, a duster & Carrying: \textit{nothing} \\
Wearing: a shirt, a sword & Wearing: \textit{nothing} \\
        \midrule
        \multicolumn{2}{l}{\textbf{{Setting:}} It's spare and humble. A small cookpot sits over a fire, and a few books sit on a shelf. A candle} \\
        \multicolumn{2}{l}{sits in the middle of a table where there are a few plates and silverware set out in preparation for} \\
        \multicolumn{2}{l}{people to eat.} \\

        \bottomrule

        \multicolumn{2}{l}{\textbf{Servant:} Hello jester, busy work day today?} \\
        \multicolumn{2}{l}{\textbf{Court jester:} Yes like always, i wish the King appreciated it more.} \\
        \multicolumn{2}{l}{\textbf{Servant:} Tell me about it, all they do is order me around and I can't do anything about it.} \\
        \multicolumn{2}{l}{\textbf{Court jester:} Would you like to grab something to eat with me?} \\
        \multicolumn{2}{l}{\textit{~~Action:} get plate} \\
        \multicolumn{2}{l}{\textbf{Servant:} Yes, I haven't eaten in a few days! What is on the menu?} \\
        \multicolumn{2}{l}{\textbf{Court jester:} It looks like fish soup! My favorite!} \\
        \multicolumn{2}{l}{\textbf{Servant:} Better than nothing, that's for sure!} \\
        \multicolumn{2}{l}{\textbf{Court jester:} I have been made fun of a lot lately. I wish i was born a knight or a noble instead of a jester..} \\
        \multicolumn{2}{l}{\textit{~~Action:} gesture frown} \\
        \multicolumn{2}{l}{\textbf{Servant:} It is tough luck that we were born so low on the totem pole.} \\
        \multicolumn{2}{l}{\textbf{Court jester:} I guess you can relate. Have you spent much time with our King?} \\
        \multicolumn{2}{l}{\textbf{Servant:} No, he only walks in and barks orders about once a week. Is he easily amused by you?} \\
        \multicolumn{2}{l}{\textbf{Court jester:} The only thing he likes about me is making fun of me.} \\
        \multicolumn{2}{l}{\textbf{Servant:} At least he laughs at you, he is always angry when he visits me.} \\
        \multicolumn{2}{l}{\textbf{Court jester:} Ugh, what a dispicable human being.} \\
      \bottomrule
      \end{tabular}
      \caption{Example dialogue from the {\sc LIGHT} dataset.
  \label{appendix:dialogues-fifth}}
    \end{footnotesize}
  \end{center}
\end{figure*}

\begin{figure*}[t]
  \begin{center}
    \begin{footnotesize}
      \begin{tabular}{l|l}
        \toprule
        \textbf{Persona 1: Spiders} & \textbf{Persona 2: Vulture} \\
        \midrule
        I am the Spider in the fable of the Spider and the Fly,  & I am a vulture that is familiar with death. \\
        much beloved by the children of the realm. & I enjoy watching living things take their last breathe. \\
        In the story, I am a kind-hearted spider, not a mean one, & I am a vital part of the ecosystem. \\
        which is why my story is considered suitable for children. &  \\
        When a fly gets caught in my sticky net, I have a choice: \\
        I can kill the fly and eat him, or I can free him and allow him \\
        to fly away. \\
        That's what I do, for I am a kind spider. \\
        \midrule
        Carrying: \textit{nothing} & Carrying: \textit{nothing} \\
Wearing: \textit{nothing} & Wearing: \textit{nothing} \\
        \midrule
        \multicolumn{2}{l}{\textbf{{Setting:}} Wispy, hot crevice that is surrounding by a bunch of skeletons. A pile of treasure sits in the} \\
        \multicolumn{2}{l}{middle. Hundreds of hungry vultures stare down upon the treasure, eager to devour any adventurer that} \\
        \multicolumn{2}{l}{draws near.} \\

        \bottomrule

        \multicolumn{2}{l}{\textbf{Spiders:} Hello vulture! It's nice to see a fellow living soul around here. I couldn't find much friendliness in}\\ 
        \multicolumn{2}{l}{these skeletons here.} \\
        \multicolumn{2}{l}{\textit{~~Action:} hug vulture} \\
        \multicolumn{2}{l}{\textbf{Vulture:} Ach, your legs are very... tickling... ahahaha, stop it!} \\
        \multicolumn{2}{l}{\textit{~~Action:} gesture laugh} \\
        \multicolumn{2}{l}{\textbf{Spiders:} Oh, I'm so sorry! I always forget that I'm so ticklish. Do you forgive me?} \\
        \multicolumn{2}{l}{\textit{~~Action:} gesture blush} \\
        \multicolumn{2}{l}{\textbf{Vulture:} Oh, well, your venomous bite took down that last adventurer quite nicely, so you're not a bad sort.}\\ \multicolumn{2}{l}{Nothing to forgive there, friend!} \\
        \multicolumn{2}{l}{\textit{~~Action:} gesture smile} \\
        \multicolumn{2}{l}{\textbf{Spiders:} Me, take down the last adventurer? I think you have the wrong idea about me. I am a friendly spider. I always free}\\ 
        \multicolumn{2}{l}{any flies that get caught in my web. I would never harm a person!} \\
        \multicolumn{2}{l}{\textbf{Vulture:} Ah, perhaps it was that scorpion over there.  I was, I admit, a bit peckish, so I might have gotten a bit forgetful} \\\multicolumn{2}{l}{amid the feasting.} \\
        \multicolumn{2}{l}{\textit{~~Action:} gesture grin} \\
        \multicolumn{2}{l}{\textbf{Spiders:} Yes, you are probably right. I tried to make friends with that scorpion but he threatened to sting me. It's sad }\\ \multicolumn{2}{l}{because I was going to give him some of the treasure I've found around here.} \\
        \multicolumn{2}{l}{\textit{~~Action:} gesture frown} \\
        \multicolumn{2}{l}{\textbf{Vulture:} Well, he looks a bit angry all the time anyways.  I mean, look at him, he's always red in the face!} \\
        \multicolumn{2}{l}{\textit{~~Action:} gesture laugh} \\
        \multicolumn{2}{l}{\textbf{Spiders:} Yes, you are quite right! But dear vulture, do you think you could help me out a bit?} \\
        \multicolumn{2}{l}{\textit{~~Action:} gesture laugh} \\
        \multicolumn{2}{l}{\textbf{Vulture:} Well, it isn't like there's much else to do.  Those gold coins are glinting in my eyes terribly, so a change of pace}\\ \multicolumn{2}{l}{would be welcome.} \\
        \multicolumn{2}{l}{\textit{~~Action:} gesture smile} \\
        \multicolumn{2}{l}{\textbf{Spiders:} Oh thank you! Can you help me on to that chair over there? I'm afraid this desert heat has taken all the energy}\\ \multicolumn{2}{l}{out of me. And I know with your power of flight, it would be easy to lift me.} \\
        \multicolumn{2}{l}{\textbf{Vulture:} Ok... just... hold still.  I wouldn't want to squish you on accident!  Here we go!} \\
        \multicolumn{2}{l}{\textit{~~Action:} hug spiders} \\
        \multicolumn{2}{l}{\textbf{Spiders:} Oh it is so nice to meet such a kind soul in such a sad dying place as this. For your kindness you will be included}\\ \multicolumn{2}{l}{in my fable, I am sure.} \\
        \multicolumn{2}{l}{\textit{~~Action:} gesture smile} \\
        \multicolumn{2}{l}{\textbf{Vulture:} Thank you?  I think.  Do you have a scribe following you about that I don't see?  I didn't know you were famous!} \\
        \multicolumn{2}{l}{\textit{~~Action:} gesture stare} \\
      \bottomrule
      \end{tabular}
      \caption{Example dialogue from the {\sc LIGHT} dataset.
  \label{appendix:dialogues-sixth}}
    \end{footnotesize}
  \end{center}
\end{figure*}

\begin{figure*}[t]
  \begin{center}
    \begin{footnotesize}
      \begin{tabular}{l|l}
        \toprule
        \textbf{Persona 1: Thief} & \textbf{Persona 2: Witch} \\
        \midrule
        I live alone in a tent in the woods. & I am a fierce witch. \\
        I steal food from the townspeople and coal from the blacksmith. & The most powerful across the realm. \\
        The village police can not find me to put me in jail. & I am feared and like to freeze people. \\
        \midrule
        Carrying: a coal & Carrying: \textit{nothing} \\
Wearing: a knife & Wearing: a hats, a dress, a Cloak, a ceremonial hat \\
        \midrule
        \multicolumn{2}{l}{\textbf{{Setting:}} An odd looking hut that sits outside that Witch's cottage. It is squat, mushy, and looks like a} \\
        \multicolumn{2}{l}{mushroom. It is be speckled with spots - black and white. The steam of the mushroom is beige. There is a} \\
        \multicolumn{2}{l}{small door that looks like it would fit a puppy through it.} \\

        \bottomrule

        \multicolumn{2}{l}{\textbf{Thief:} Hello witch, waht brings you here?} \\
        \multicolumn{2}{l}{\textbf{Witch:} Good day.  I am here to collect ingredients for my spells.} \\
        \multicolumn{2}{l}{\textbf{Thief:} Which ingredients do you seek?} \\
        \multicolumn{2}{l}{\textbf{Witch:} A black mushroom, covered in green mold. } \\
        \multicolumn{2}{l}{\textbf{Thief:} Ironic, everything looks like a muchroom around here. I can help you.} \\
        \multicolumn{2}{l}{\textbf{Witch:} You think I need help from a lowly thief? You're lucky I don't freeze you where you stand.} \\
        \multicolumn{2}{l}{\textit{~~Action:} gesture smile} \\
        \multicolumn{2}{l}{\textbf{Thief:} I can be of some help because i know exactly where that muchrooms flourishes. But i want something in return.} \\
        \multicolumn{2}{l}{\textbf{Witch:} Name your price.} \\
        \multicolumn{2}{l}{\textbf{Thief:} I wish too look different. I am wanted and i dont want them to recognize me. Can you do that?} \\
        \multicolumn{2}{l}{\textbf{Witch:} That is easy.  But it also requires a rare ingredient I don't have, tongue of raven.  You must procure that.} \\
        \multicolumn{2}{l}{\textit{~~Action:} gesture nod} \\
        \multicolumn{2}{l}{\textbf{Thief:} Interesting, have you seen any ravens nearby?} \\
        \multicolumn{2}{l}{\textbf{Witch:} They fly over the abandoned church. If you are clever enough to catch one I can change your looks.} \\
        \multicolumn{2}{l}{\textbf{Thief:} I think i have an idea on how to catch one. Will you coem with me to catch one? It iwll only take a moment.} \\
        \multicolumn{2}{l}{\textbf{Witch:} Get my mushroom first.  I will not change you until I get my ingredients.} \\
        \multicolumn{2}{l}{\textit{~~Action:} remove ceremonial hat} \\
      \bottomrule
      \end{tabular}
      \caption{Example dialogue from the {\sc LIGHT} dataset.
  \label{appendix:dialogues-last}}
    \end{footnotesize}
  \end{center}
\end{figure*}

\end{document}